%% file: latex/main.tex
\documentclass[11pt]{article}

\usepackage[preprint]{acl}

\usepackage{times}
\usepackage{latexsym}

\usepackage[T1]{fontenc}

\usepackage[utf8]{inputenc}

\usepackage{microtype}

\usepackage{inconsolata}

\usepackage{graphicx}

\usepackage{amsmath}
\usepackage{amssymb}
\usepackage{enumitem}

\usepackage{graphicx}     
\usepackage{booktabs}     
\usepackage{array}        
\usepackage{multirow}    
\usepackage{makecell}   
\usepackage{xcolor}
\usepackage{color, colortbl}
\usepackage[most]{tcolorbox}
\usepackage{amsfonts}
\tcbuselibrary{theorems}
\usepackage{ulem} 
\usepackage{adjustbox}
\usepackage{hyperref}     
\usepackage{url}          
\usepackage[table]{xcolor}
\usepackage{xcolor}       
\usepackage{subfigure}
\usepackage{subcaption}

\newtcolorbox{mybox}[2][]
  {colback = black!5!white, colframe = black!75!black, fonttitle = \bfseries,
    colbacktitle = black!100!black, enhanced, before upper={\fontsize{10}{11}\obeyspaces\obeylines\selectfont}, fontupper=\selectfont,
    attach boxed title to top left={yshift=-2.2mm,xshift=2mm},
    title=#2,#1}
\title{What Makes Low-Bit Quantization-Aware Training Work for Reasoning LLMs? A Systematic Study}

\author{\textbf{Keyu Lv$^{1*}$},
\textbf{Manyi Zhang$^{2}$}\thanks{Equal contribution. \ $^{\#}$Corresponding authors.},
\textbf{Xiaobo Xia$^{3}$},
\textbf{Jingchen Ni$^{1}$},
\\\textbf{Shannan Yan$^{1}$,}
\textbf{Xianzhi Yu$^{2}$,}
\textbf{Lu Hou$^{2}$,}
\textbf{Chun Yuan$^{1\#}$,}
\textbf{Haoli Bai$^{2\#}$}
\\
$^{1}$Shenzhen International Graduate School, Tsinghua University \\
$^{2}$Huawei Technologies \ \ $^{3}$National University of Singapore \\
\small\texttt{lvky24@mails.tsinghua.edu.cn} \ \
\small\texttt{yuanc@sz.tsinghua.edu.cn} \ \
\small\texttt{baihaoli@huawei.com} \\
}

\begin{document}
\maketitle
\begin{abstract}
Reasoning models excel at complex tasks such as coding and mathematics, yet their inference is often slow and token-inefficient. To improve the inference efficiency, post-training quantization~(PTQ) usually comes with the cost of large accuracy drops, especially for reasoning tasks under low-bit settings. In this study, we present a systematic empirical study of quantization-aware training (QAT) for reasoning models. Our key findings include: (1) Knowledge distillation is a robust objective for reasoning models trained via either supervised fine-tuning or reinforcement learning; (2) PTQ provides a strong initialization for QAT, improving accuracy while reducing training cost; (3) Reinforcement learning remains feasible for quantized models given a viable cold start and yields additional gains; and (4) Aligning the PTQ calibration domain with the QAT training domain accelerates convergence and often improves the final accuracy. Finally, we consolidate these findings into an optimized workflow (Reasoning-QAT), and show that it consistently outperforms state-of-the-art PTQ methods across multiple LLM backbones and reasoning datasets. For instance, on Qwen3-0.6B, it surpasses GPTQ by 44.53\% on MATH-500 and consistently recovers performance in the 2-bit regime. 
\end{abstract}

\input{sections/Intro}
\input{sections/preliminary}
\input{sections/benchmark}

\input{sections/method}

\input{sections/conclusion}
\bibliography{custom}

\appendix

\input{sections/appendix}

\end{document}

%% file: sections/Intro.tex
\section{Introduction}
Recent large language models (LLMs)~\citep{jaech2024openai, guo2025deepseek, team2025kimi} with enhanced reasoning capabilities have achieved remarkable progress in domains such as coding and mathematics. However, this progress comes with a deployment bottleneck: reasoning-focused inference is often slow and token-inefficient, resulting in substantial inference overhead~\citep{qu2025survey}. Quantization is a widely used technique to accelerate LLM inference~\citep{li2024evaluating,liu2024intactkv,li2025gptaq, ma2024affinequant, lin2024duquant, li2025kvtuner, yangattentionpredictor, zhang2026benchmarking}, yet prior studies~\citep{li2025quantization,srivastava2025towards,liu2025quantization, wang2025bitnet} show that under extreme low-bit settings (e.g., 3-bit or 2-bit weight-only quantization), post-training quantization (PTQ) can trigger severe accuracy degradation on reasoning benchmarks. We corroborate this phenomenon by comparing quantized LLMs on both non-reasoning and reasoning tasks (Figure~\ref{fig:nonr_vs_r}): while 4-bit group-wise weight quantization (group size 128) is near-lossless across tasks, 3-bit variants incur large drops, and the degradation is remarkably larger on reasoning tasks than on non-reasoning ones.

Quantization-aware training (QAT)~\citep{tailor2020degree,nagel2022overcoming, bondarenko2024low, jeon2024l4q, qin2024accurate} is an appealing alternative to recover the performance drop by simulating low-precision inference during training. While QAT has demonstrated effectiveness for general-purpose LLMs~\citep{liu2023llm, chen2024efficientqat}, it remains unclear whether these benefits extend to reasoning-focused models, and it is non-trivial to address. For instance, our preliminary attempts show that applying reinforcement learning (RL) directly on a severely degraded quantized policy often fails to explore valid reasoning trajectories under quantization noise. This motivates this study: \textit{what are the key factors that lead to the success of QAT with reasoning models?}

In this study, we present a systematic study of quantization-aware training (QAT) for reasoning models. We investigate the following critical factors: 1) the choice of training objective, e.g., supervised fine-tuning~(SFT) vs.\ knowledge distillation~(KD)~\cite{hinton2015distilling}; 2) the role of PTQ initialization; 3) the integration of RL with QAT; and 4) the choice of QAT training data. We study two representative quantization settings---3-bit and 2-bit weight-only quantization with group size 128,  covering two major reasoning training paradigms: (i) supervised fine-tuning (SFT), represented by DeepSeek-R1-Qwen-Distill-1.5B~\citep{guo2025deepseek}, and (ii) reinforcement learning (RL), represented by Qwen3-0.6B and Qwen3-4B~\citep{yang2025qwen3}. We evaluate on a diverse suite of reasoning benchmarks, including AIME-120, MATH-500~\citep{lightman2023let}, GSM8K~\citep{cobbe2021gsm8k}, GPQA-Diamond~\citep{rein2024gpqa}, and LiveCodeBench~\citep{jain2024livecodebench}. We summarize our findings as follows:

\begin{itemize}[leftmargin=2ex]
\setlength{\itemsep}{2pt}
\item \textbf{Training Objective}~(\S\ref{sec:rq1}): Knowledge distillation (KD)~\citep{hinton2015distilling} is the preferred objective for QAT, which can effectively boost reasoning models trained by either SFT or RL. 
\item \textbf{PTQ Initialization}~(\S\ref{sec:rq2}): PTQ provides a strong  initialization that effectively saves the training cost and stabilizes the QAT training,  especially in early stages.
\item \textbf{QAT with Reinforcement Learning}~(\S\ref{sec:rq3}): With KD training as the cold start, QAT with RL can deliver further performance gains.
\item \textbf{Choice of QAT Data}~(\S\ref{sec:rq4}): Aligning the domain of QAT dataset with the calibration set by PTQ is beneficial for QAT training, i.e., it yields a faster and more stable training process.
\end{itemize}

Finally, based on the findings above, we optimize the QAT workflow for reasoning models, termed as \textit{Reasoning-QAT}. Specifically, Reasoning-QAT is structured in the following way: \textit{PTQ-based initialization $\rightarrow$ KD-based recovery $\rightarrow$ Cold-start RL}. Our empirical results show that this workflow consistently outperforms state-of-the-art PTQ and QAT baselines across multiple backbones and reasoning benchmarks. For example, on Qwen3-0.6B, it surpasses GPTQ~\citep{frantar2022gptq} by 44.53\% on MATH-500 under 3-bit quantization; meanwhile, on DeepSeek-R1-Distill-Qwen-1.5B, it outperforms representative QAT baselines by up to 4.75\% on average.
We hope our research provides valuable guidance toward better quantization methods for reasoning models.

%% file: sections/preliminary.tex
\section{Preliminaries and Research Questions}

\subsection{Post-training Quantization for Reasoning Models}

\paragraph{Background and Notations.} 
Quantization has been a popular approach for the compression and acceleration of LLMs. Given model parameters $\mathbf{W}$ stored in bfloat16, quantization converts $\mathbf{W}$ into low-bit integer representations $\mathbf{W}_{int}$, i.e., 
\begin{equation}\label{eq:quant}
    \mathbf{W}_{int} = \mathrm{clip}(\lfloor \frac{\mathbf{W}}{s} \rceil + z, Q_{min}, Q_{max}),
\end{equation}

where $\mathrm{clip}(\cdot,\  Q_{min},\  Q_{max})$ clips values into the range $[Q_{min}, Q_{max}]$, $s$ is the scaling factor and $z$ is the zero point. For $N$-bit symmetric quantization, $s = \frac{\max(|\mathbf{W}|)}{2^{N-1}-1}$ and $z=0$. For asymmetric quantization, $s = \frac{max(\mathbf{W})-min(\mathbf{W})}{2^{N}-1}$, $z=\lfloor \frac{-min(\mathbf{W})}{s}\rceil$. 

For weight quantization, the low bit quantized weights $\mathbf{W}_{int}$ in the forward pass are dequantized to $\hat{\mathbf{W}} = s\cdot(\mathbf{W_{int}} - z)$ for the following operations. For completeness, in weight-activation quantization, both weights and activations are stored as low-bit integers and computed with integer kernels, which can further reduce compute beyond memory savings.

\paragraph{Post-training quantization incurs a large performance drop on reasoning models.}
Most prior work on LLM quantization focuses on post-training quantization (PTQ)~\citep{frantar2022optq,lin2023awq,ashkboos2024quarot,sun2024flatquant, liu2025flexquant}, where the model is directly quantized without training. PTQ is usually fast and easy to implement, with satisfactory performance on many general natural language tasks. However, recent studies~\citep{liu2025quantization} show that quantized reasoning models can exhibit large performance drops, particularly on challenging reasoning benchmarks.

To further validate this, we compare PTQ-quantized LLMs on both non-reasoning and reasoning benchmarks. From Figure~\ref{fig:nonr_vs_r}, it can be found that for DeepSeek-R1-Distill-Qwen-1.5B (abbr.\ R1-Qwen-1.5B for simplicity in the following text), the performance degradation on reasoning tasks (e.g., 11.67\%$\downarrow$ on AIME-120 and 12.80\%$\downarrow$ on MATH-500) is much larger than that on non-reasoning tasks (e.g., 1.03\%$\downarrow$ on Winogrande, 3.13\%$\downarrow$ on Hellaswag). Similar observations can be found for Qwen3-4B. Therefore, under extreme low-bit settings, PTQ alone is often insufficient to preserve reasoning performance, motivating training-time adaptation such as QAT.

\input{figs/reason_vs_nonreason/reason_vs_nonreason}

\subsection{Quantization-aware Training for Reasoning Models}
\label{sec:qat}

To mitigate the performance degradation of PTQ, QAT is a commonly used alternative. QAT simulates low-precision inference during training, allowing model weights to adapt to quantization effects. In the forward pass, QAT inserts fake quantization operations to obtain quantized weights $\hat{\mathbf{W}}$ or activations $\hat{\mathbf{X}}$ in each linear layer, and optimizes the training objective $\mathcal{L}(\hat{\mathbf{W}})$. In the backward pass, since quantization is non-differentiable, the straight-through estimator~(STE) is typically used to pass gradients to the original weights $\mathbf{W}$, e.g., $\frac{\partial \mathcal{L}}{\partial \mathbf{W}} = \frac{\partial \mathcal{L}}{\partial \hat{\mathbf{W}}}\cdot \mathbf{1}({Q_{min}\leq \mathbf{W}/s \leq Q_{max}})$. 

While QAT has been shown effective for general-purpose LLMs~\citep{liu2023llm,chen2024efficientqat}, how these benefits extend to reasoning models remains unclear. In this study, we aim to investigate the following four research questions~(RQs):

\begin{tcolorbox}[colback=gray!10,colframe=gray!35!black, left=1ex]
\begin{description}    
    \item[RQ1.] Which training objective is most suitable for QAT on reasoning models?
    \item[RQ2.] What improves the training efficiency of QAT under low-bit settings?
    \item[RQ3.] How does QAT interact with RL (e.g., GRPO) in the low-bit regime?
    \item[RQ4.] How does the choice of QAT training data affect quantized reasoning performance?
\end{description}
\end{tcolorbox}

\paragraph{Training Paradigm and Objectives (RQ1 \& RQ3).}
The optimal training methodology for QAT on reasoning models is unclear. Standard QAT often reuses the cross-entropy objective from pre-training or instruction fine-tuning~\citep{liu2025paretoq,lee2024improving}. In contrast, many reasoning models are trained either by supervised fine-tuning with teacher-generated data (often paired with knowledge distillation)~\citep{guo2025deepseek} or by reinforcement learning~\citep{guo2025deepseek,team2025kimi,yang2025qwen3}. How to integrate the training objectives (e.g., SFT vs.\ KD) as prerequisites for stable low-bit training, and the combination of QAT with RL, remain unexplored.

\paragraph{Training Efficiency and Overhead (RQ2).}
Severe accuracy degradation under extreme low-bit quantization often requires substantial additional training to recover reasoning performance. This can make QAT time- and compute-intensive, hindering practical deployment when the retraining budget is limited. Identifying strategies to improve sample and time efficiency of QAT is therefore a major practical concern.

\paragraph{Data Strategy (RQ4).}
Beyond the training objective, the selection of data for QAT itself is a critical factor. The impact of QAT training data (including its domain, quality, and alignment with calibration data) on the reasoning performance of the quantized model is not well characterized. A systematic analysis is needed to guide efficient and effective data curation.

%% file: figs/reason_vs_nonreason/reason_vs_nonreason.tex
\begin{figure*}[t]
\centering
\includegraphics[width=0.4\linewidth]{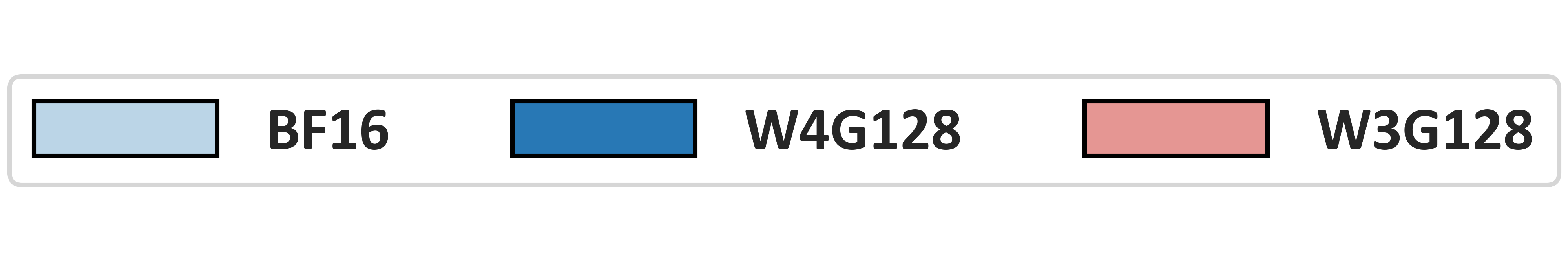}
\par
\vspace{-2ex}
\subfigure[DeepSeek-R1-Distill-Qwen-1.5B.]{\includegraphics[width=.48\textwidth]{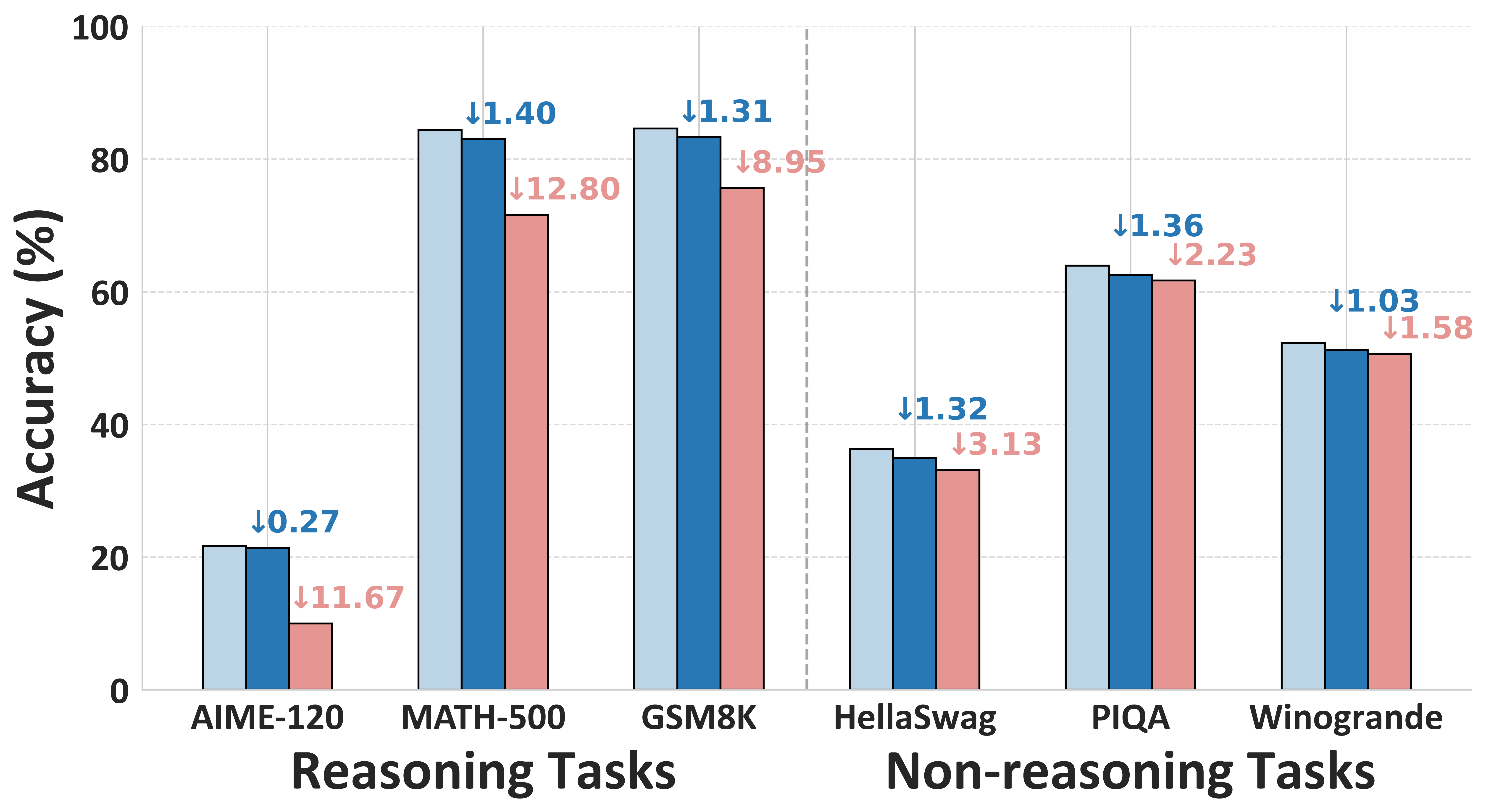}}
\subfigure[Qwen3-4B.]{\includegraphics[width=.48\textwidth]{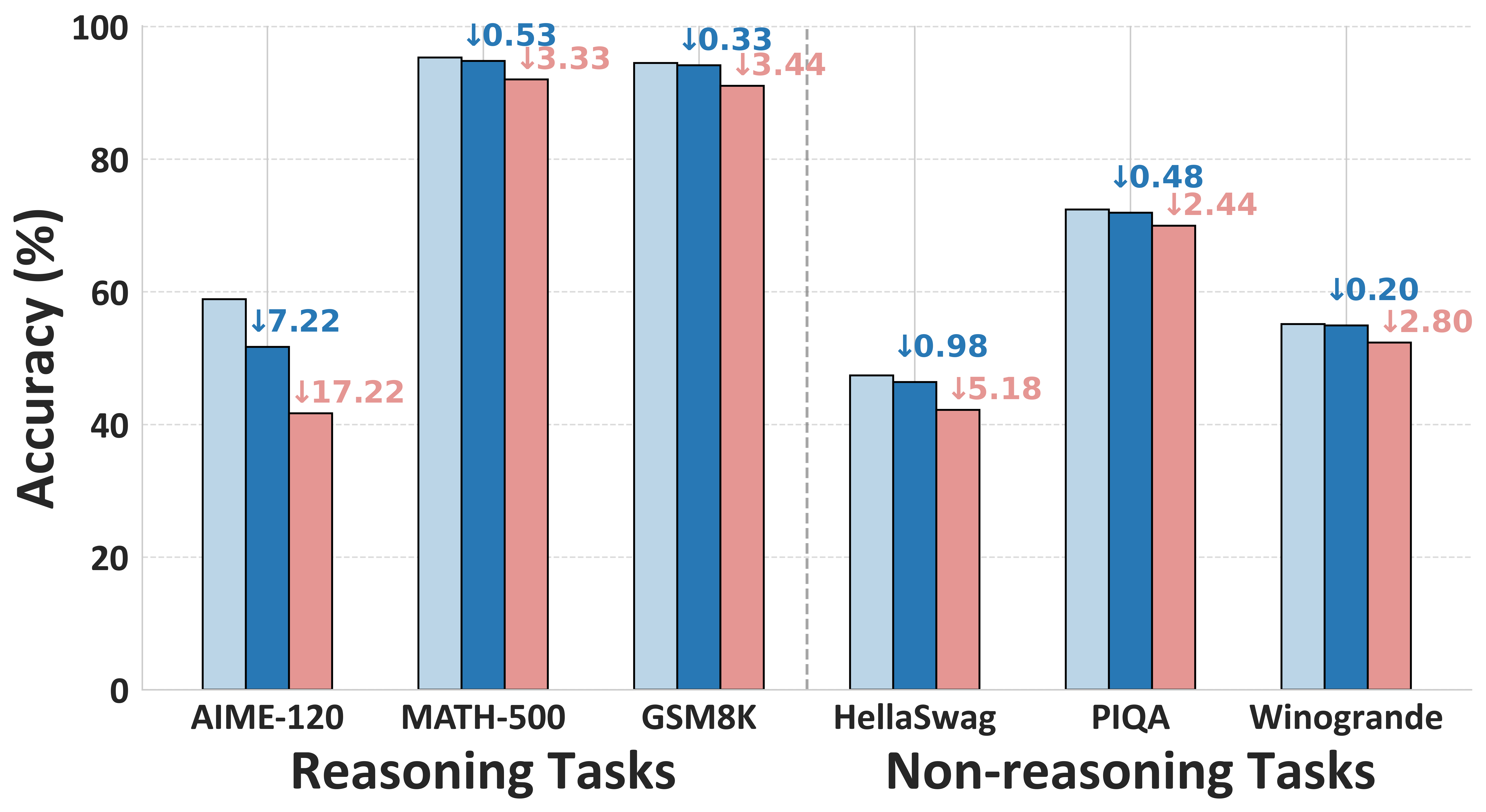}}
\vspace{-1ex}
\caption{The performance degradation by post-training quantization on reasoning and non-reasoning tasks. We adopt GPTQ with 3-bit weight only quantization with group size 128, and the results are based on DeepSeek-R1-Distill-Qwen-1.5B and Qwen-4B. }
\label{fig:nonr_vs_r}
\end{figure*}

%% file: sections/benchmark.tex
\section{A Systematic Study of QAT for Reasoning Models}
\label{sec:benchmark}
In this study, we conduct a systematic empirical study of quantization-aware training for reasoning models, aiming to address the research questions posed in \S\ref{sec:qat}.

\subsection{Setups}
\label{setup}

\paragraph{Quantization Settings.} We quantize all linear layers of the model, excluding the token embedding and \texttt{lm\_head} layers. Our primary focus is on 3-bit and 2-bit group-wise weight-only quantization with a group size of 128 (W3G128 and W2G128). For initialization, we consider two commonly used schemes: a symmetric round-to-nearest (RTN) baseline and an asymmetric GPTQ initialization. For completeness, we also report results under a joint 4-bit weight and 4-bit activation (W4A4) setting in Appendix~\ref{extensive_wa_exp}.

\paragraph{Models and Dataset.} We evaluate two categories of reasoning models. For SFT-based reasoning models, we adopt DeepSeek-R1-Distill-Qwen-1.5B~\citep{guo2025deepseek}. For RL-trained reasoning models, we use Qwen3-0.6B and Qwen3-4B~\citep{yang2025qwen3}, the two recent competitive open-source LLMs.
For the choice of training dataset, 
our study includes two training phases with distinct data configurations. In the initial fine-tuning phase (SFT and KD), we use OpenR1-Math~\citep{openr1} (94k problems in the default subset) for the weight-only low-bit setting. In the subsequent reinforcement learning (RL) phase, we also use the OpenR1-Math dataset. This design allows us to examine the effect of domain consistency between the QAT training data and the calibration data (\S\ref{sec:rq4}). For completeness, the data configuration for the W4A4 setting is reported in Appendix~\ref{extensive_wa_exp}.

\paragraph{Evaluation Benchmarks.} We assess quantized models across training paradigms on a suite of reasoning benchmarks: (1) three mathematical reasoning benchmarks sorted by difficulty: AIME-120 (120 problems from AIME 2022--2025), MATH-500~\citep{lightman2023let}, and GSM8K~\citep{cobbe2021gsm8k}; (2) LiveCodeBench~\citep{jain2024livecodebench} for code generation; and (3) GPQA-Diamond~\citep{rein2024gpqa} for graduate-level science question answering. All evaluations are conducted with LightEval~\citep{lighteval} using the vLLM~\citep{kwon2023efficient} backend. We set temperature to 0.6, top-$p$ to 0.95, and the maximum number of generated tokens to 32,768. We report average scores over three random seeds.

\paragraph{Training Implementations.} We implement and evaluate three training objectives in our study. For supervised fine-tuning (SFT), we use the standard cross-entropy loss. For knowledge distillation (KD), the quantized model serves as the student and is trained to match the output distribution of the full-precision teacher by minimizing KL divergence. For reinforcement learning (RL), we employ Group Relative Policy Optimization (GRPO)~\citep{shao2024deepseekmath}. Further hyperparameter details are provided in Appendix~\ref{appendix:train_details}.

\subsection{QAT Training Objectives: SFT or KD?}
\label{sec:rq1}

We investigate QAT objectives (SFT vs. KD) for reasoning models trained via supervised fine-tuning~(e.g., R1-Qwen-1.5B) or reinforcement learning~(e.g., Qwen3-4B). Adopting W3G128 weight-only quantization, Table~\ref{tab:kd_sft} compares the performance recovery of KD and SFT. Key observations include: 1) KD outperforms SFT on both model types. Specifically, SFT suffers average accuracy drops of 10.51\%$\downarrow$ and 21.40\%$\downarrow$ on R1-Qwen-1.5B and Qwen3-4B, respectively, whereas KD limits drops to 8.06\%$\downarrow$ and 9.26\%$\downarrow$; 2) KD exhibits consistent synergy across paradigms, with similar drops for both models~(8.06\% vs. 9.26\%). In contrast, SFT causes a moderate drop on R1-Qwen-1.5B but a more severe one~(21.40\%$\downarrow$) on Qwen3-4B. We hypothesize that KD provides smoother signals by aligning output distributions, preserving uncertainty structure better than hard-label SFT. \textit{We therefore recommend KD over SFT for QAT due to its superior performance and robustness across training paradigms.}

\begin{mybox}[colback=gray!10]{Findings (RQ1)}
    \begin{itemize}[leftmargin=2ex]
      \item KD is preferable to SFT as the QAT objective for reasoning models, and is more consistent across both SFT-trained and RL-trained reasoning models.
    \end{itemize}
\end{mybox}

\begin{table}[t]
\centering
\resizebox{0.95\linewidth}{!}{
\renewcommand\arraystretch{1.4} 
\begin{tabular}{c|c|c|ccccl}
\toprule
\textbf{Model} & \textbf{Setting} & \textbf{Method} & \textbf{AIME120} & \textbf{MATH-500} & \textbf{GSM8K} & \textbf{AVG} & \textbf{Drop$\downarrow$} \\
\midrule
\multirowcell{4}{\centering\rotatebox{90}{R1-Qwen-1.5B}}
&BF16 & - & 21.67 & 84.4 & 84.61 & 63.56 & -- \\
\cline{2-8}
&\multirow{3}{*}{W3G128} & RTN & 0.83 & 15.00 & 15.39 & 10.41 & 53.15  $\downarrow$\\
                            && SFT & 10.00 & 73.60 & 75.54 & 53.05 & 10.51 $\downarrow$ \\
                            &&\cellcolor{red!10} KD  & \cellcolor{red!10} 14.44 &\cellcolor{red!10} 76.20 &\cellcolor{red!10} 75.87 &\cellcolor{red!10} 55.50 & \cellcolor{red!10} 8.06 $\downarrow$ \\
\midrule
\multirowcell{4}{\centering\rotatebox{90}{Qwen3-4B}}
&BF16 & - & 58.89 & 95.33 & 94.49 & 82.90 & -- \\
\cline{2-8}
&\multirow{3}{*}{W3G128} & RTN & 0.00 & 1.40 & 0.99 & 0.80 & 82.10 $\downarrow$ \\
                           & & SFT & 14.44 & 81.80 & 88.25 & 61.50 & 21.40 $\downarrow$ \\
                           & &\cellcolor{red!10} KD  & \cellcolor{red!10} 37.50&\cellcolor{red!10} 92.00 &\cellcolor{red!10} 91.43 &\cellcolor{red!10} 73.64 & \cellcolor{red!10} 9.26$\downarrow$ \\
\bottomrule
\end{tabular}
}
\vspace{-1ex}

\caption{Objective choice for low-bit QAT. KD yields smaller accuracy drops than SFT on both an SFT-trained model (R1-Qwen-1.5B) and an RL-trained model (Qwen3-4B) under W3G128, while SFT degrades much more on the RL-trained model.}

\label{tab:kd_sft}
\end{table}

\subsection{Training Efficiency of QAT}
\label{sec:rq2}

\textbf{Initializing QAT with PTQ.} In previous work, it has been standard practice to initialize QAT from a pretrained full-precision model~\citep{liu2023llm,du2024bitdistiller}. Here, we systematically investigate how PTQ-based initializations affect the convergence and accuracy of QAT. Specifically, we employ GPTQ~\citep{frantar2022gptq} for initialization, using weights that have been adjusted via Hessian-based compensation prior to the quantization process. As shown in Figure~\ref{fig:PTQ_init}(a)-(b), we compare the test accuracy and training loss of RTN+KD and GPTQ+KD on the MATH-500 benchmark using the R1-Qwen-1.5B model. Using the GPTQ-initialized weights, GPTQ+KD starts from a higher starting point (higher test accuracy and lower loss). Furthermore, GPTQ+KD consistently outperforms RTN+KD and exhibits a faster convergence rate within the same number of training steps. This is likely because PTQ provides a more accurate starting point than RTN, which minimizes the initial performance gap and mitigates optimization difficulties. Therefore, \textit{PTQ acts as a strong initialization to improve the training efficiency of QAT.}

\input{figs/efficiency_vis/efficiency_vis}

\paragraph{Training Efficiency: KD vs.\ SFT.} In addition to studying the initialization of quantized models, we further compare the training efficiency of KD versus SFT, as shown in Figure~\ref{fig:kd_vs_sft}. \textit{The results show that KD consistently achieves higher accuracy than SFT and also converges faster.}

\begin{mybox}[colback=gray!10]{Findings (RQ2)}
    \begin{itemize}[leftmargin=2ex]
      \item PTQ acts as a strong initialization to minimize the initial performance gap, while KD converges faster than SFT, collectively improving training efficiency and accuracy.
    \end{itemize}
\end{mybox}

\subsection{QAT with Reinforcement Learning}
\label{sec:rq3}

While Reinforcement Learning~(RL) enhances LLM reasoning, its role in QAT remains underexplored. Our results identify a critical prerequisite: \textit{RL requires proper initialization to avoid collapse}. We compare \textit{zero-RL QAT} (applying RL directly to RTN-quantized models) and \textit{cold-start RL QAT} (initializing with a KD-tuned model). Using GRPO~\citep{guo2025deepseek} with correctness rewards, Table~\ref{tab:rl_condition} shows that zero-RL QAT collapses completely, whereas the cold-start setting improves accuracy by approximately 46\%. RTN-based models with severe quantization errors fail to sample valid outputs, preventing effective reward generation. Conversely, KD recovers sampling capability, ensuring sufficient reward density. We conclude: \textit{RL alone cannot rescue heavily quantized models but drives improvement given a KD cold start.}

\begin{table}[t]
\centering
\resizebox{\linewidth}{!}{%
\begin{tabular}{ccc|ccc|c}
\toprule
\textbf{RTN} & \textbf{KD} & \textbf{GRPO} & \textbf{AIME120} & \textbf{MATH-500} & \textbf{GSM8K} & \textbf{AVG} \\
\cline{1-7}
- & - & - & 21.67 & 84.40 & 84.61 & 63.56 \\
\hline
\checkmark & - & - & 0.83 & 15.00 & 15.39 & 10.41 \\
\checkmark & - & \checkmark & 1.67 & 15.33 & 15.52 & 10.84 \\
\checkmark & \checkmark & \checkmark & 14.44 & 78.00 & 77.93 & 56.79 \\
\bottomrule
\end{tabular}%
}
\vspace{-1ex}
\caption{RL under low-bit quantization (R1-Qwen-1.5B, W3G128). KD cold-start is necessary for effective RL.}
\label{tab:rl_condition}
\end{table}

\paragraph{Roles of RL under low-bit QAT.}

Figure~\ref{fig:rl_curve} demonstrates the indispensable roles of RL in QAT. First, as shown in Figure~\ref{fig:rl_curve}(a), RL simultaneously increases reward and reduces excessive response length. This prevents the model from exploiting response length to increase rewards, guiding it toward high-quality outputs. Second, in Figure~\ref{fig:rl_curve}(b), RL drives a decrease in entropy, which reduces prediction randomness and enforces deterministic outputs. This avoids collapse and ensures stable convergence despite quantization errors. Lastly, in Figure~\ref{fig:rl_curve}(c), RL improves test accuracy while reducing response length, enhancing generalization without unnecessary verbosity. We hypothesize that RL works by reducing entropy, helping the model resist quantization noise.

\input{figs/rl_vis/rl_vis}

\begin{mybox}[colback=gray!10]{Findings (RQ3)}
    \begin{itemize}[leftmargin=2ex]
      \item KD cold-start serves as a practical prerequisite for RL under low-bit QAT, enabling subsequent RL to enhance both model performance and efficiency.
    \end{itemize}
\end{mybox}

\subsection{The Choice of QAT Training Dataset}
\label{sec:rq4}

\input{figs/data_consistency/data_consistency}
The choice of the QAT training dataset remains an open challenge, specifically regarding how domain differences influence optimization dynamics and final performance. We compare two datasets: Wikitext2~(natural language) and OpenR1-Math~(reasoning-based math). Following \S\ref{sec:rq2}, we initialize QAT models using PTQ. Calibration uses either Wikitext2 or NuminaMath-1.5, where the latter is closely aligned with OpenR1-Math\footnote{OpenR1-Math consists of reasoning traces generated by DeepSeek R1 for problems from NuminaMath 1.5. See \url{https://huggingface.co/datasets/open-r1/OpenR1-Math-220k}.}. We then conduct KD for QAT on both datasets, yielding four combinations. Figure~\ref{fig:data_consistency} shows test accuracy curves, revealing three observations:
First, domain alignment accelerates convergence: when the calibration and the training domains match, the curve converges much earlier (e.g., \texttt{numina+openr1}), whereas mismatched combinations stabilize slowly, changing beyond \texttt{5k+} steps. Second, domain mismatch causes harmful readjustment. Specifically, \texttt{numina+wiki} starts with strong initialization but suffers a sharp accuracy drop after switching to Wikitext2, recovering slowly without returning to the initial level. This implies the training domain significantly alters the PTQ-calibrated solution, making cross-domain adaptation unstable. Third, reasoning-domain data is critical for final performance: configurations using reasoning data (especially \texttt{numina+openr1}) achieve the highest accuracy, while general text (\texttt{wiki+wiki}) leads to substantially lower accuracy. Overall, these results suggest aligning PTQ calibration and QAT training domains, and confirm that reasoning data is crucial for recovering low-bit performance.

\begin{mybox}[colback=gray!10]{Findings (RQ4)}
\begin{itemize}[leftmargin=2ex]
  \item Aligning the PTQ calibration domain with the QAT training domain accelerates convergence, whereas utilizing reasoning-domain data is key to achieving a high final reasoning performance.
\end{itemize}
\end{mybox}

%% file: figs/efficiency_vis/efficiency_vis.tex
\begin{figure*}
    \centering
    \begin{minipage}[t]{0.64\textwidth}
        \centering
        \subfigure[Test Accuracy]{\includegraphics[width=.48\textwidth]{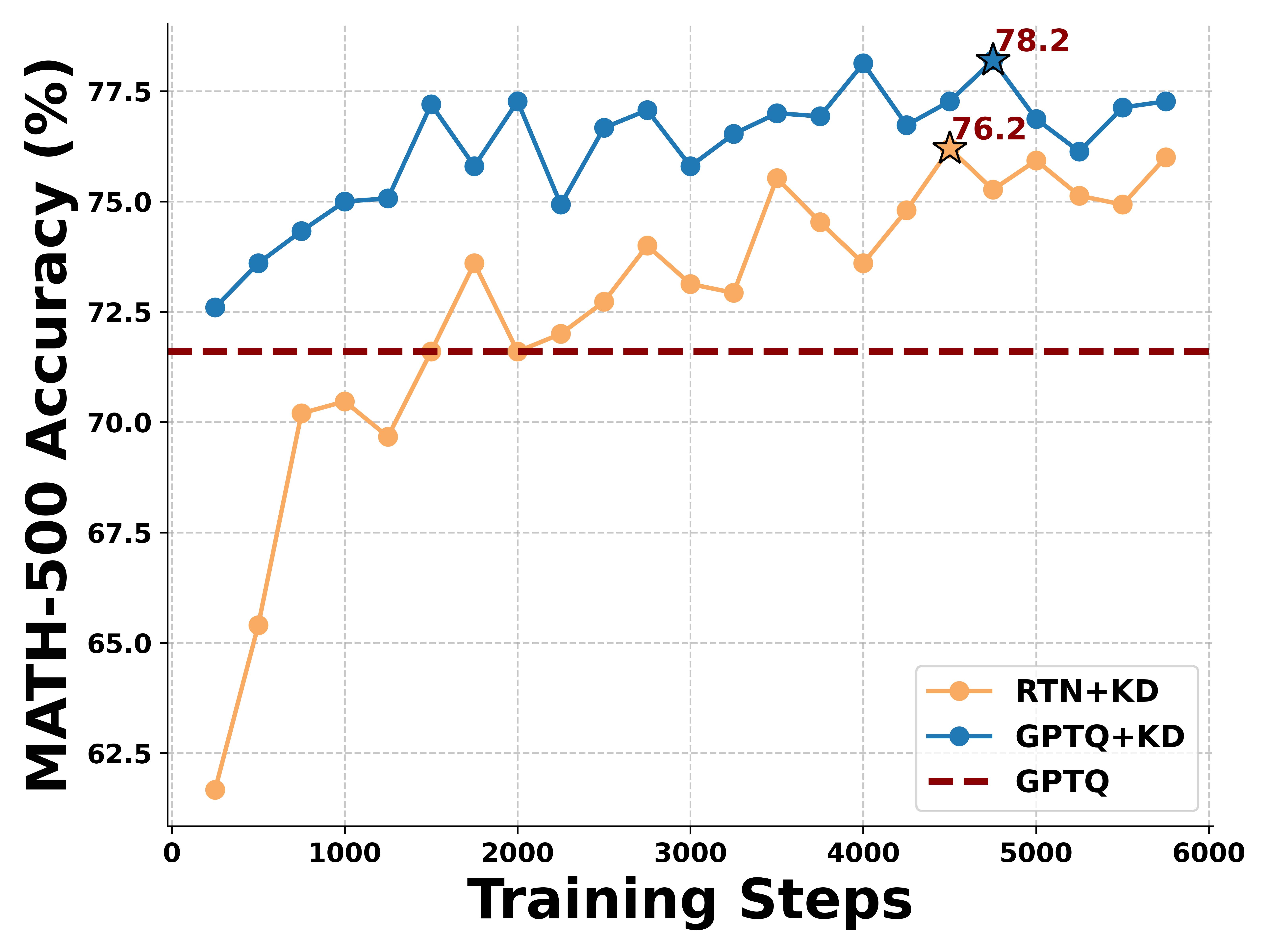}}
        \hspace{-1.5pt}
        \subfigure[Loss Value]{\includegraphics[width=.48\textwidth]{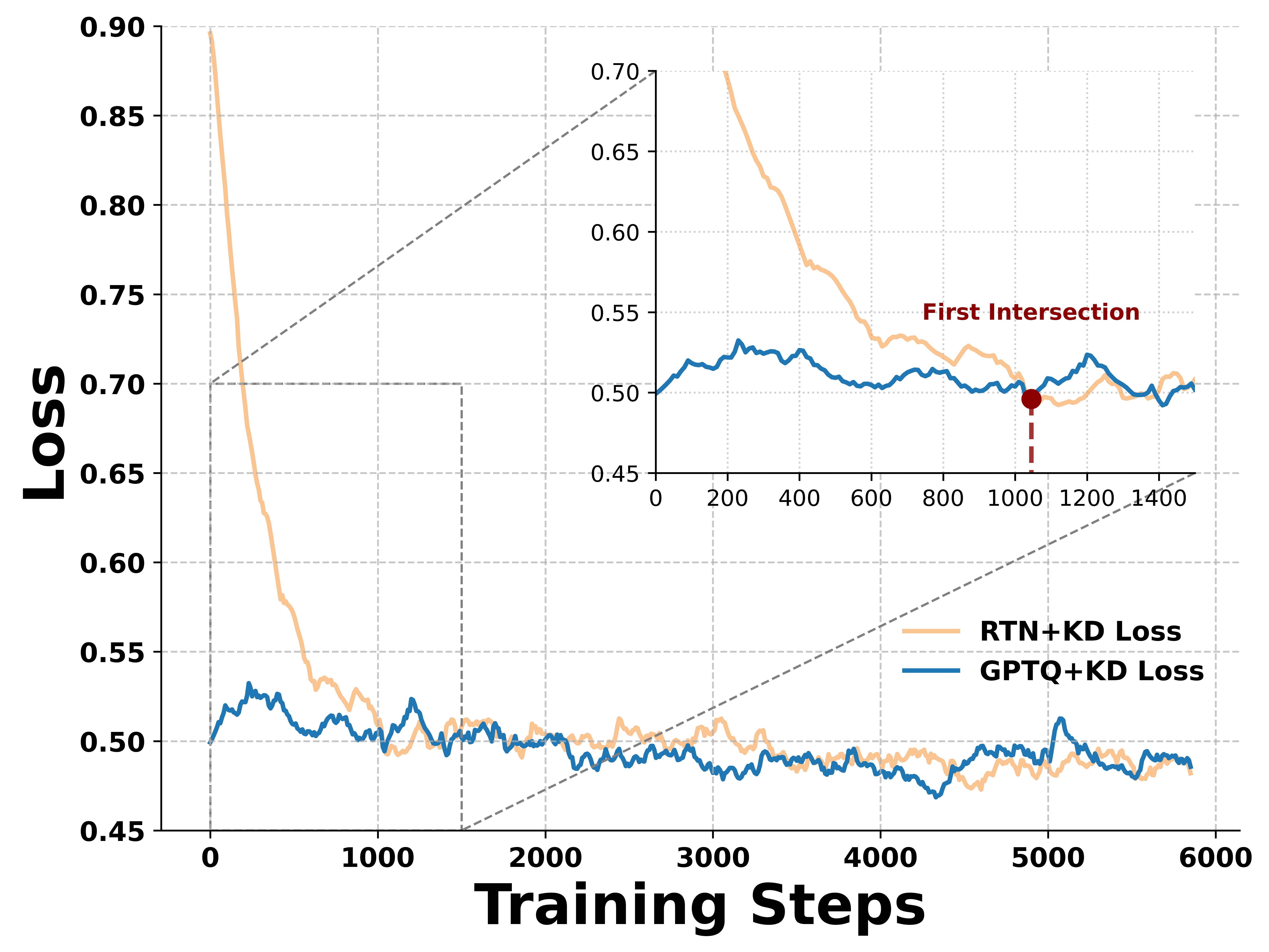}}
        \caption{PTQ initialization improves QAT efficiency (RTN+KD vs.\ GPTQ+KD on MATH-500): (a) accuracy, (b) loss.}
        \label{fig:PTQ_init}
    \end{minipage}
    \hspace{1.5pt}
    \begin{minipage}[t]{0.32\textwidth}
        \centering
        \vspace{5pt}
        \subfigure[Test Accuracy]{
            \includegraphics[width=0.96\linewidth]{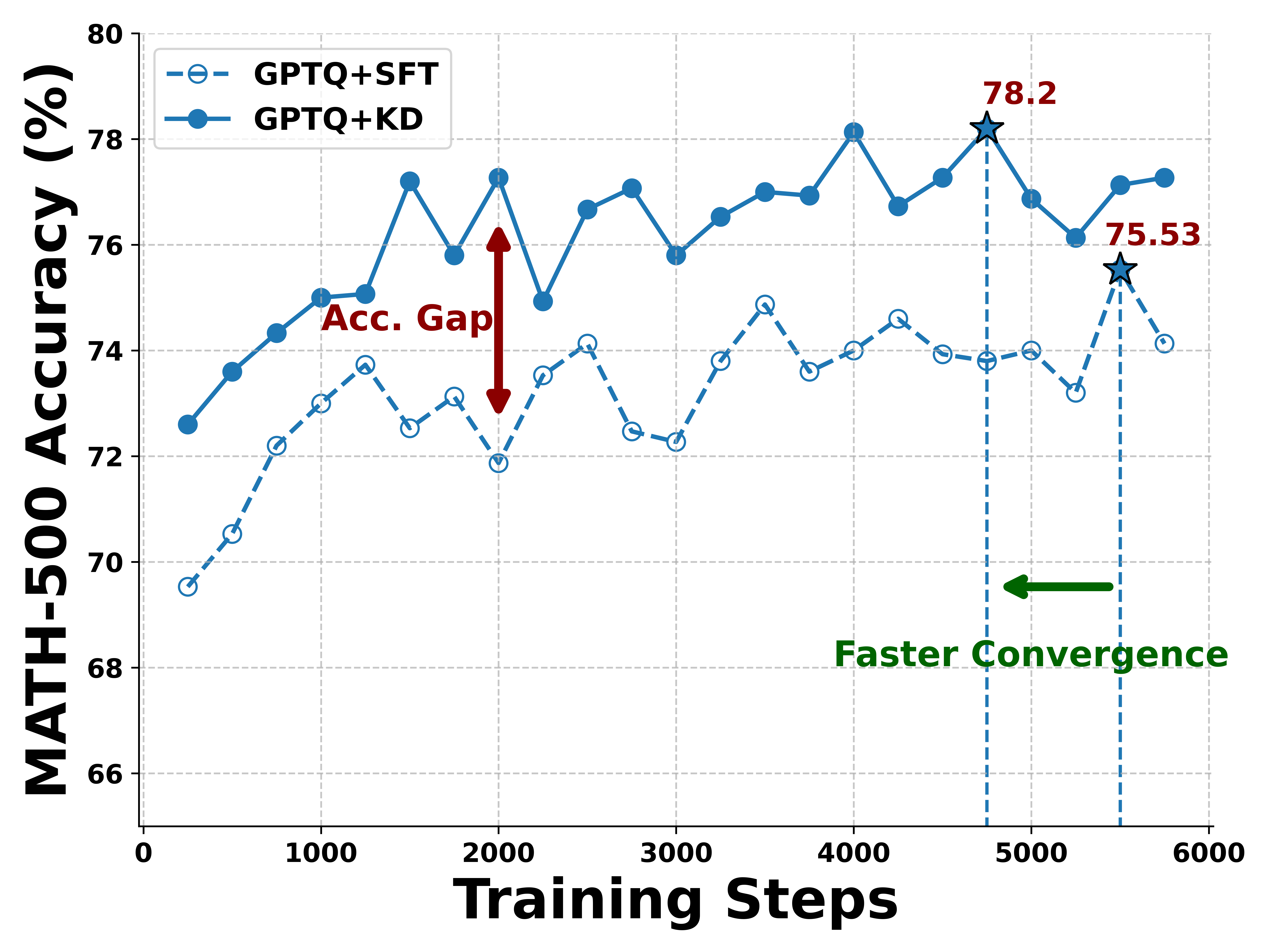}
        }

        \caption{KD vs.\ SFT under GPTQ initialization on MATH-500.}
        \label{fig:kd_vs_sft}
    \end{minipage}
\end{figure*}

%% file: figs/rl_vis/rl_vis.tex
\begin{figure*}[t]
\centering
\subfigure[]{\includegraphics[width=.32\textwidth]{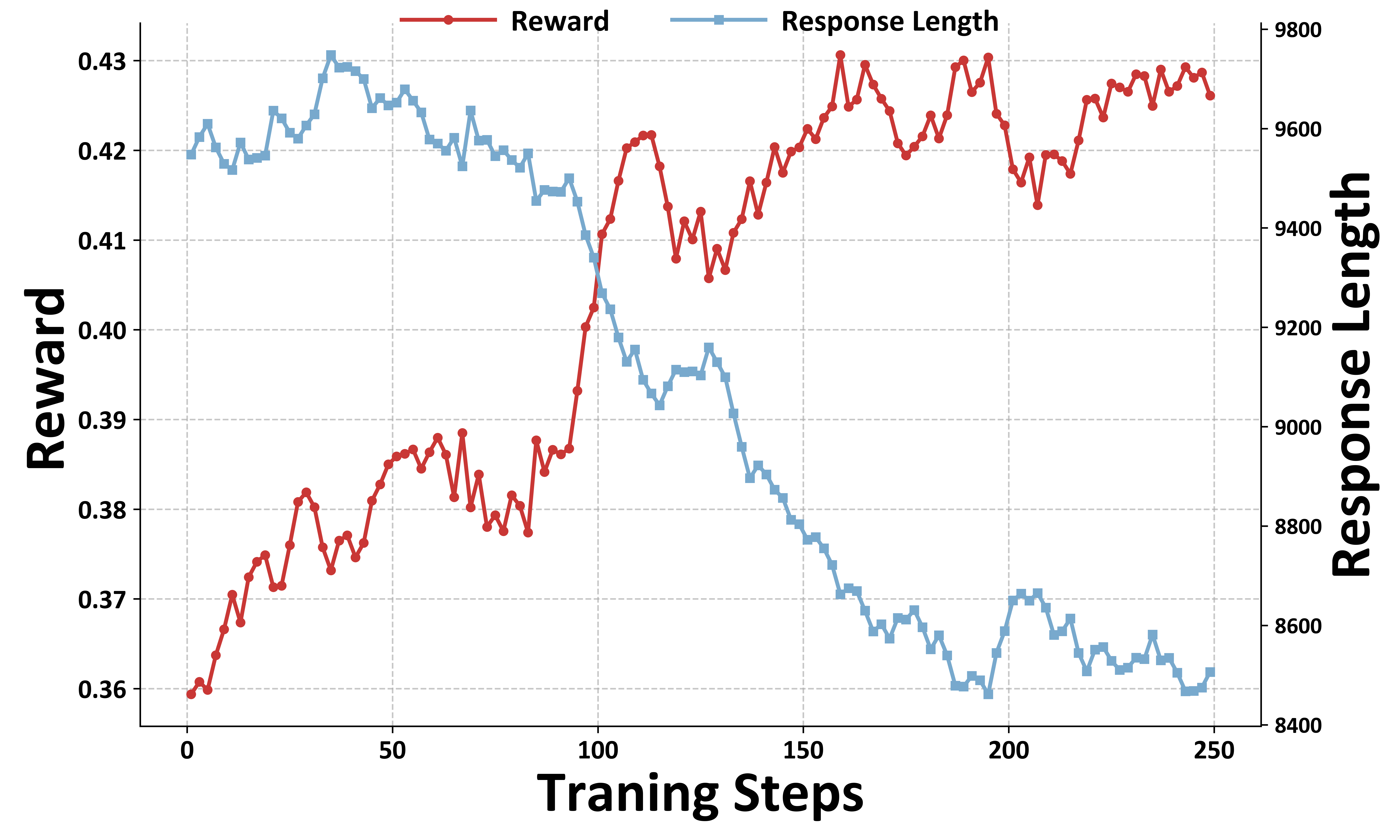}}
\subfigure[]{\includegraphics[width=.32\textwidth]{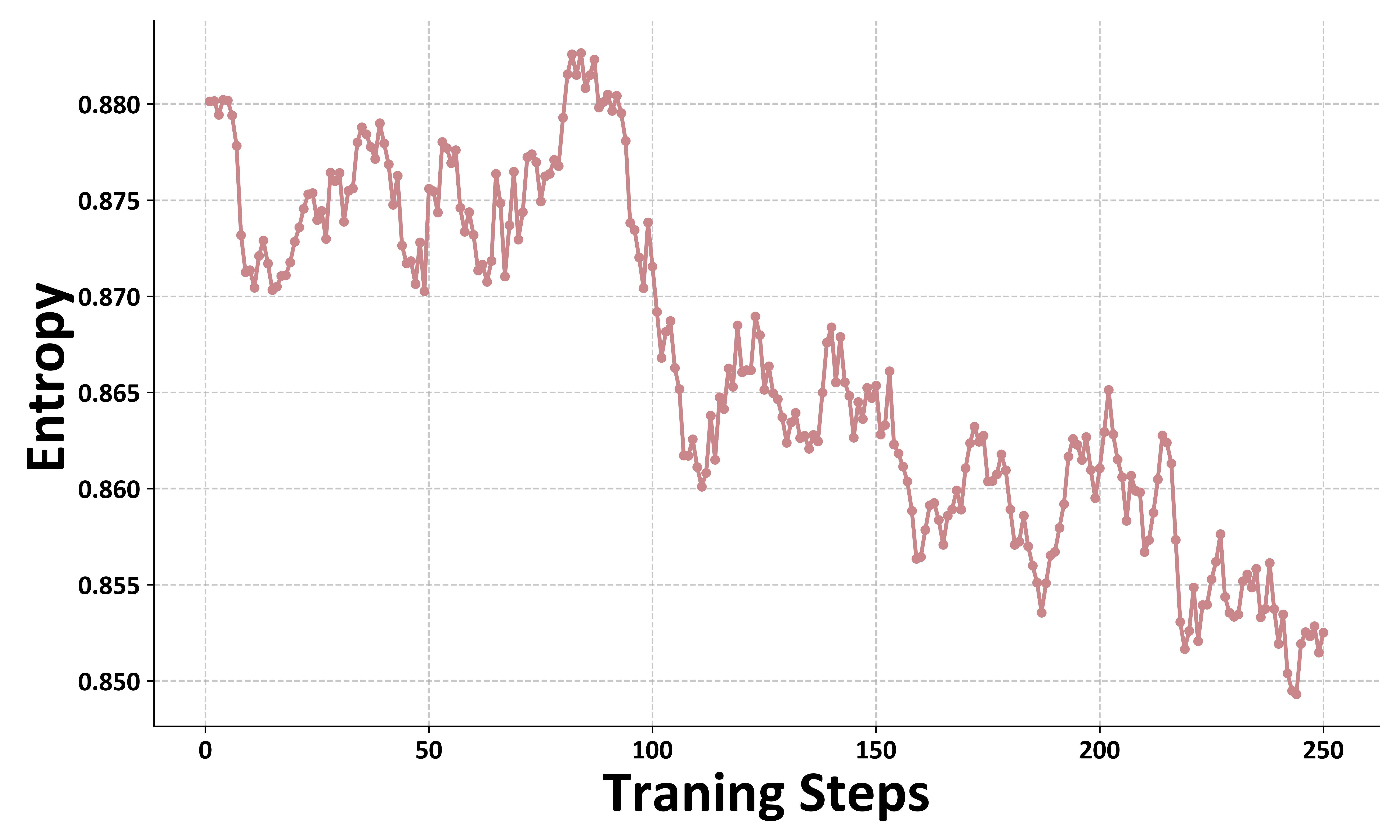}}
\subfigure[]{\includegraphics[width=.32\textwidth]{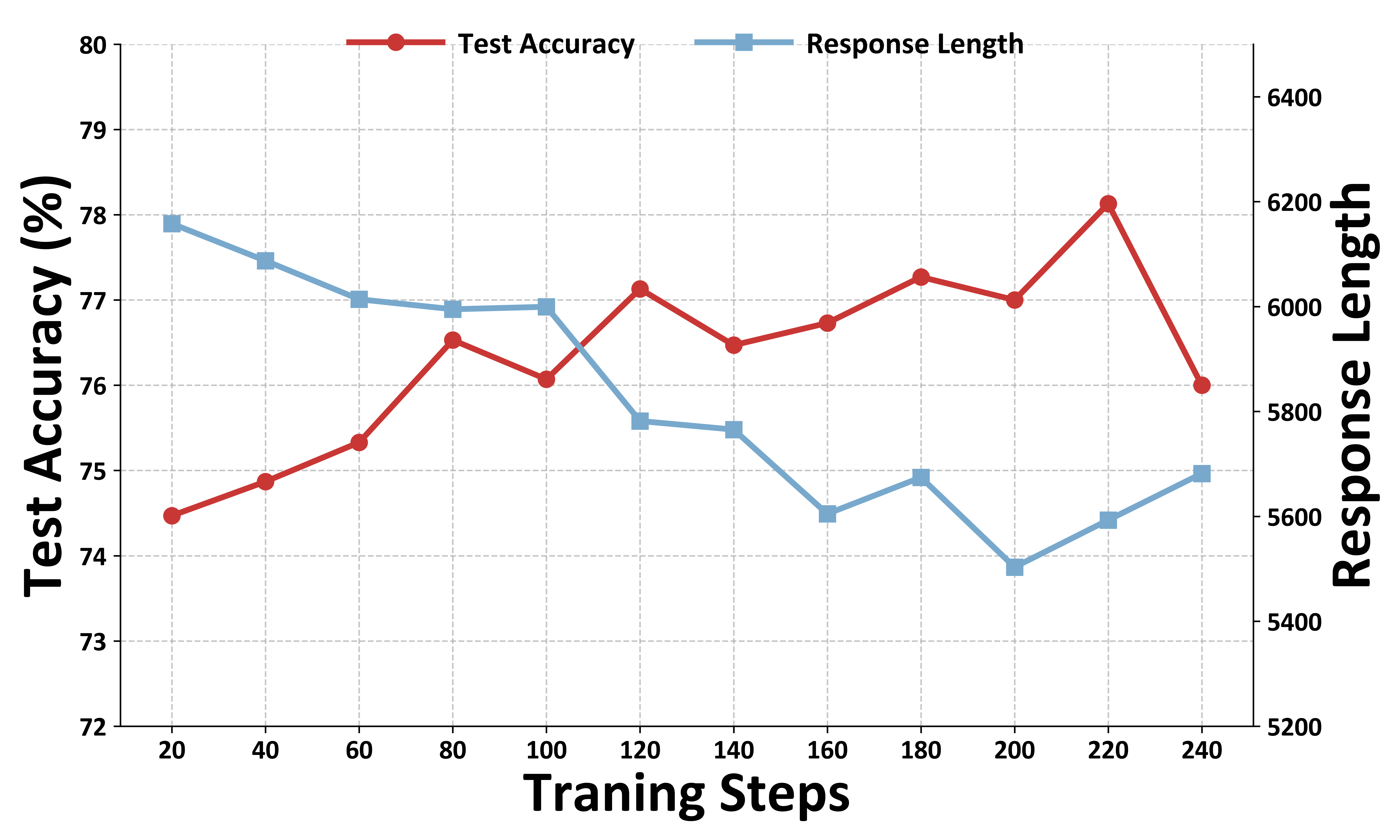}}
\vspace{-1ex}
\caption{RL effects after KD cold-start (W3G128). (a) reward$\uparrow$ with length$\downarrow$, (b) entropy$\downarrow$, (c) test accuracy$\uparrow$ with length$\downarrow$.}
\label{fig:rl_curve}
\end{figure*}

%% file: figs/data_consistency/data_consistency.tex
\begin{figure}
    \centering
    \includegraphics[width=0.8\linewidth]{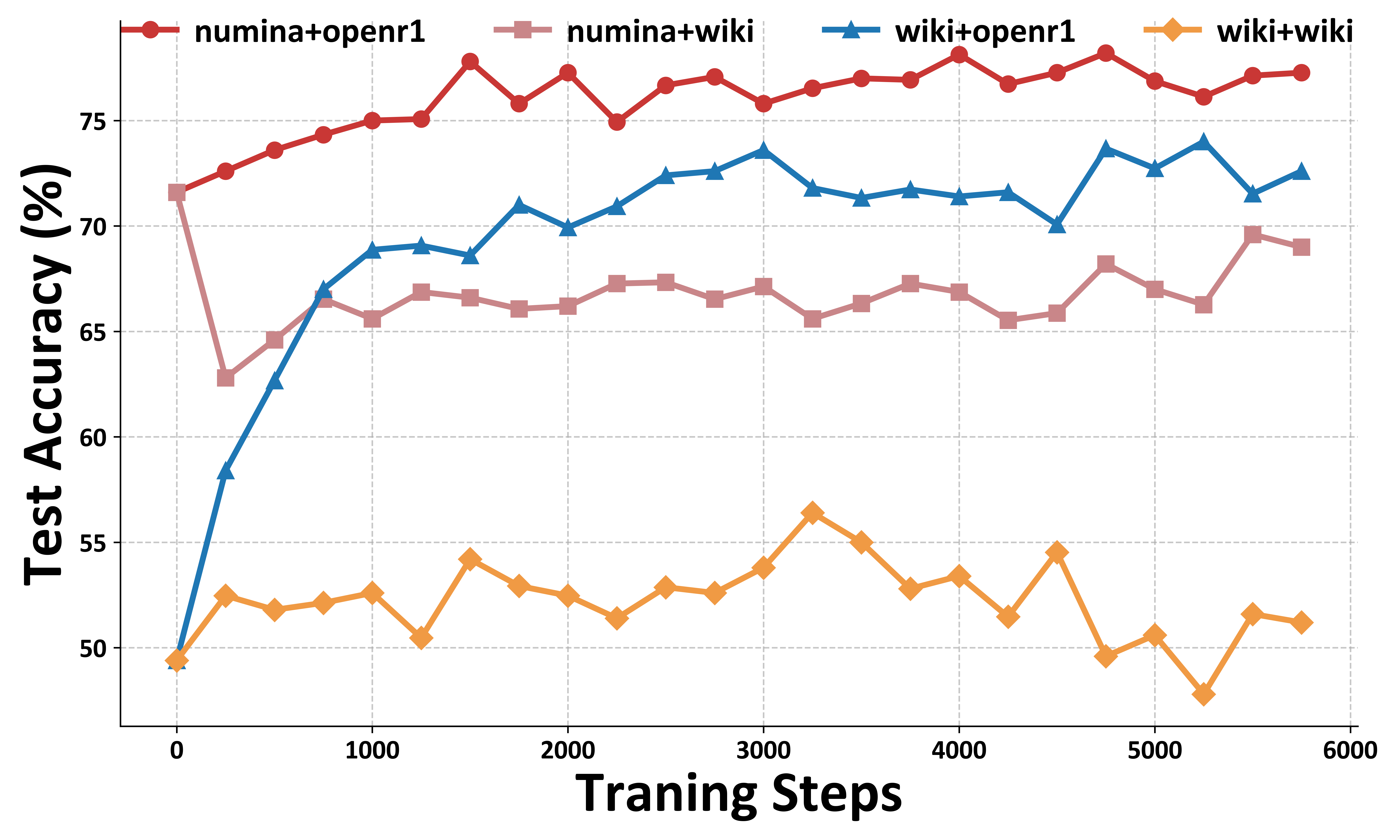}
    \caption{Comparison with different combinations of PTQ and QAT data domain. We present test accuracy curves on MATH-500 for the R1-Qwen-1.5B model under W3G128 setting.}
    \vspace{-2ex}
    \label{fig:data_consistency}
\end{figure}

%% file: sections/method.tex
\section{Reasoning-QAT: The Optimized Workflow}
Based on the observations in \S\ref{sec:benchmark}, we provide the optimized workflow in Figure~\ref{fig:pipeline}, which includes three key steps to guide practical applications and support downstream usage. Note that this section does not propose a new algorithm; rather, it validates our findings and offers feasible guidance, showing that satisfying the identified dependencies is sufficient for stable low-bit training. For completeness, we include an additional weight-activation setting in Appendix~\ref{extensive_wa_exp}.

\subsection{The Reasoning-QAT Workflow}
\input{figs/pipeline/pipeline}

We construct this reference workflow by strictly satisfying the dependencies identified in RQ1--RQ3:

\begin{itemize}[leftmargin=2ex]
\setlength{\itemsep}{2pt}
\item \textbf{PTQ-based Initialization.} Motivated by \S\ref{sec:rq2}, we rectify the latent weights with PTQ techniques as the initial state for QAT. While the QAT model still retains continuous weights, this initialization strategy improves its tolerance to quantization and provides a better starting point for subsequent training.
\item \textbf{Knowledge Distillation.} We then perform knowledge distillation from the original full-precision model. Guided by the findings in \S\ref{sec:rq1}, this step fine-tunes the QAT model to align its output distribution with that of the full-precision model. After that, the distilled model not only recovers from the quantization-induced degradation, but also serves as a stable cold-start actor for RL.

\item \textbf{Cold-start RL.} Following the prerequisites discussed in \S\ref{sec:rq3}, we apply RL on top of the knowledge-distilled model from Step~2. Here, we employ GRPO~\citep{guo2025deepseek} as the RL paradigm. This cold start design avoids the collapse issue observed when directly using RL on heavily quantized models, while utilizing the stabilized initialization to ensure reliable optimization. During this stage, RL progressively enhances the reasoning capability of the quantized model, driving more deterministic outputs and reducing randomness. 
\end{itemize}

\subsection{Empirical Evaluations}

\begin{table*}[t]
\centering
\resizebox{0.9\linewidth}{!}{%
\begin{tabular}{p{1.2cm}|c|c|rrrrr|r|r}
    \toprule
    \textbf{Model} & \textbf{W-Bits} &  \textbf{Methods} & \textbf{AIME-120} & \textbf{MATH-500} & \textbf{GSM8K} & \textbf{\makecell{GPQA-\\Diamond}} & \textbf{\makecell{LiveCode-\\Bench}} & \textbf{Avg.} & \textbf{Drop $\downarrow$} \\
    \midrule
    \multirowcell{8}{\centering\rotatebox{90}{Qwen3-0.6B}}
    & BF16 &  - & 11.11 & 74.00 & 79.00 & 28.45 & 12.94 & 41.10 & - \\
    \cline{2-10}
    & \multirowcell{4}{W3G128}
        & RTN & 0.00 & 0.80 & 0.30 & 24.24 & 0.00 & 5.07 & -36.03 \\
    & &  GPTQ & 0.00 & 13.27 & 23.33 & 26.43 & 0.00 & 12.61 & -28.49 \\
    & &  AWQ & 0.00 & 5.20 & 10.01 & 26.77 & 0.00 & 8.40 & -32.70 \\
    & &  \cellcolor{red!10} Reasoning-QAT & \cellcolor{red!10} 3.89 & \cellcolor{red!10} 57.80 & \cellcolor{red!10} 67.02 & \cellcolor{red!10} 27.78 & \cellcolor{red!10} 1.87 & \cellcolor{red!10} 31.67 & \cellcolor{red!10} -9.43 \\
    \cline{2-10}
    & \multirowcell{3}{W2G128}
    &  GPTQ & 0.00 & 0.60 & 0.13 & 0.84 & 0.00 & 0.31 & -40.79 \\
    & &  AWQ & 0.00 & 0.00 & 0.00 & 23.91 & 0.00 & 4.78 & -36.32 \\
    & & \cellcolor{red!10} Reasoning-QAT & \cellcolor{red!10} 0.56 & \cellcolor{red!10} 20.20 & \cellcolor{red!10} 30.33 & \cellcolor{red!10} 25.25 &\cellcolor{red!10} 0.00 & \cellcolor{red!10} 15.27 &\cellcolor{red!10} -25.83 \\
    \cline{1-10}

    \multirowcell{8}{\centering\rotatebox{90}{R1-Qwen-1.5B}}
    & BF16 &  - & 21.67 & 84.40 & 84.61 & 36.87 & 16.04 & 48.72 & - \\
    \cline{2-10}
    & \multirowcell{4}{W3G128}
        & RTN & 0.83 & 15.00 & 15.39 & 19.19 & 0.00 & 10.08 & -38.64 \\
    & &  GPTQ & 10.00 & 71.60 & 75.66 & 23.74 & 9.33 & 38.07 & -10.65 \\
    & &  AWQ & 3.33 & 48.80 & 65.81 & 37.88 & 4.85 & 32.13 & -16.58 \\
    & &  \cellcolor{red!10} Reasoning-QAT & \cellcolor{red!10} 16.39 & \cellcolor{red!10} 79.80 &\cellcolor{red!10} 79.35 &\cellcolor{red!10} 30.30 &\cellcolor{red!10} 10.45 &\cellcolor{red!10} 43.26 &\cellcolor{red!10} -5.46 \\
    \cline{2-10}
    & \multirowcell{3}{W2G128}
    &  GPTQ & 0.00 & 3.67 & 2.86 & 21.89 & 0.00 & 5.68 & -43.04 \\
    & &  AWQ & 0.00 & 0.00 & 0.00 & 25.08 & 0.00 & 5.02 & -43.70 \\
    & & \cellcolor{red!10} Reasoning-QAT & \cellcolor{red!10} 5.15 &\cellcolor{red!10} 55.00 &\cellcolor{red!10} 55.02 &\cellcolor{red!10} 25.75 &\cellcolor{red!10} 0.00 &\cellcolor{red!10} 28.18 &\cellcolor{red!10} -20.54 \\
    \cline{1-10}   
    \multirowcell{8}{\centering\rotatebox{90}{Qwen3-4B}}
    & BF16 &  - & 58.89 & 95.33 & 94.49 & 56.06 & 48.38 & 70.63 & - \\
    \cline{2-10}
    & \multirowcell{4}{W3G128}
        & RTN & 0.00 & 1.40 & 0.99 & 10.60 & 0.00 & 2.60 & -68.03 \\
    & &  GPTQ & 34.17 & 90.07 & 91.74 & 38.05 & 20.77 & 54.96 & -15.67 \\
    & &  AWQ & 25.00 & 87.00 & 90.07 & 37.88 & 19.03 & 51.80 & -18.83 \\
    & &\cellcolor{red!10}  Reasoning-QAT &\cellcolor{red!10} 41.11 &\cellcolor{red!10} 93.47 &\cellcolor{red!10} 93.48 &\cellcolor{red!10} 45.79 &\cellcolor{red!10} 38.06 &\cellcolor{red!10} 62.38 &\cellcolor{red!10} -8.25 \\
    \cline{2-10}
    & \multirowcell{3}{W2G128}
    &  GPTQ & 0.00 & 4.80 & 5.28 & 20.70 & 0.00 & 6.16 & -64.47 \\
    & & AWQ & 0.00 & 0.00 & 0.00 & 25.59 & 0.00 & 5.12 & -65.61 \\
    & &\cellcolor{red!10}  Reasoning-QAT &\cellcolor{red!10} 22.78 & \cellcolor{red!10} 78.27 &\cellcolor{red!10} 82.96
    &\cellcolor{red!10} 25.42 & \cellcolor{red!10} 0.00 &\cellcolor{red!10} 41.89 &\cellcolor{red!10} -28.74 \\
    \bottomrule
\end{tabular}%
}

\caption{Main results of the reference configuration (Reasoning-QAT) versus representative PTQ baselines across models and reasoning benchmarks. }
\label{tab:complex_example}
\end{table*}
\paragraph{Comparison with Representative PTQ Baselines.}
We compare Reasoning-QAT against commonly used PTQ baselines under two settings: 3-bit weight-only (W3G128) and 2-bit weight-only (W2G128) quantization.

\paragraph{3-Bit Weight-only Quantization.}
Table~\ref{tab:complex_example} shows that PTQ baselines (e.g., RTN/GPTQ/AWQ) can incur severe degradation on reasoning benchmarks in the 3-bit regime. In contrast, Reasoning-QAT consistently recovers a substantial portion of the lost accuracy across model scales.
For example, on Qwen3-0.6B the average score increases from 12.61\% (GPTQ) to 31.67\% (Reasoning-QAT). On R1-Qwen-1.5B and Qwen3-4B, the remaining gap to BF16 is reduced to -5.46\% and -8.25\%, respectively, outperforming the PTQ baselines under the same setting. Overall, these results suggest that when PTQ alone struggles at low bits, the dependency-satisfying workflow provides a strong empirical recovery baseline.

\paragraph{2-Bit Weight-only Quantization.}
The 2-bit setting (W2G128) is substantially more challenging, and PTQ baselines largely collapse on math-centric benchmarks. In contrast, Reasoning-QAT recovers non-trivial reasoning accuracy, with particularly large gains on math datasets.
For instance, on R1-Qwen-1.5B, MATH-500 improves from 3.67\% (GPTQ) to 55.00\%. On Qwen3-4B, MATH-500 improves from 4.80\% (GPTQ) to 78.27\% and AIME-120 reaches 22.78\%. This highlights the necessity of QAT under extremely low-bit quantization.

\paragraph{Comparison with Representative QAT Baselines.}
We additionally compare Reasoning-QAT with two representative QAT methods developed for general-purpose LLMs, EfficientQAT~\citep{chen2024efficientqat} and BitDistiller~\citep{du2024bitdistiller}. Because these methods were not originally reported on reasoning benchmarks, we reproduce them under the same protocol (R1-Qwen-1.5B, W3G128, OpenR1-Math) for a fair comparison.

\begin{table}[t]
\centering

\resizebox{\linewidth}{!}{
\begin{tabular}{l|ccc|c}
\toprule
\textbf{Method} & \textbf{AIME120} & \textbf{MATH-500} & \textbf{GSM8K} & \textbf{AVG} \\
\midrule
Standard SFT-QAT & 10.00 & 73.60 & 75.54 & 53.05 \\
EfficientQAT~\citep{chen2024efficientqat}  & 10.83 & 74.20 & 76.26 & 53.76 \\
BitDistiller~\citep{du2024bitdistiller} & 14.72 & 78.00 & 78.46 & 57.06 \\
\midrule
\textbf{Reasoning-QAT} & \textbf{16.39} & \textbf{79.80} & \textbf{79.35} & \textbf{58.51} \\
\bottomrule
\end{tabular}
}
\caption{Comparison with representative QAT baselines on R1-Qwen-1.5B (W3G128) under the same protocol (OpenR1-Math). EfficientQAT and BitDistiller are reproduced for this setting. The best result is shown in bold.} 
\vspace{-2ex}
\label{tab:sota_qat_comparison}
\end{table}

Table~\ref{tab:sota_qat_comparison} shows that Reasoning-QAT achieves the best average accuracy (58.51\%) among the reproduced QAT baselines. The gap is moderate but consistent (e.g., +4.75\% over EfficientQAT and +1.45\% over BitDistiller), supporting our central observation that low-bit QAT for reasoning models: KD provides a stronger recovery objective, and once a viable policy is established, RL can further improve model performance.

\subsection{Analysis of Workflow Components}
\label{sec:coherence_check}

In this section, we clarify the efficacy of each workflow component, which are PTQ initialization, KD, and GRPO. We specifically assess W3G128 on R1-Qwen-1.5B model shown in Table~\ref{tab:final_abla} (detailed analysis in Appendix~\ref{complete_combination}).

\paragraph{The Effect of PTQ Initialization. }
GPTQ yields a markedly better low-bit starting point than RTN (AVG: 52.42\% vs.\ 10.41\%), which is also consistent with the findings in \S\ref{sec:rq2}.

\paragraph{The Effect of KD. }
Under the same GPTQ initialization, KD improves over SFT from 55.27\% to 56.45\%, indicating that distillation remains beneficial even when starting from a strong PTQ baseline. The result aligns with \S\ref{sec:rq1} that KD is the preferred objective for low-bit QAT on reasoning models.

\paragraph{Further Improvement by GRPO.}
Finally, adding GRPO on top of KD yields additional gains: 56.45\% $\rightarrow$ 58.51\% under GPTQ. This is consistent with \S\ref{sec:rq3}: RL is effective after KD cold start has established a viable low-bit policy, at which point RL can further improve performance. Overall, the workflow (GPTQ+KD+GRPO) reaches 58.51\% on AVG, which narrows the gap to the BF16 upper bound, i.e., 63.56\%.

\begin{table}[t]
\centering
\resizebox{\linewidth}{!}{
\renewcommand\arraystretch{1.15}
\begin{tabular}{l|ccc|c}
\toprule
\textbf{Configuration} & \textbf{AIME120} & \textbf{MATH-500} & \textbf{GSM8K} & \textbf{AVG} \\
\midrule
BF16  & 21.67 & 84.40 & 84.61 & 63.56 \\
\midrule
RTN & 0.83 & 15.00 & 15.39 & 10.41 \\
GPTQ (PTQ init) & 10.00 & 71.60 & 75.66 & 52.42 \\
GPTQ + SFT & 14.17 & 75.53 & 76.12 & 55.27 \\
GPTQ + KD & 13.89 & 78.20 & 77.26 & 56.45 \\
\textbf{GPTQ + KD + GRPO} & \textbf{16.39} & \textbf{79.80} & \textbf{79.35} & \textbf{58.51} \\
\bottomrule
\end{tabular}
}
\vspace{-1ex}
\caption{Ablation study on R1-Qwen-1.5B (W3G128), full results are reported in Appendix~\ref{complete_combination} (Table~\ref{tab:final_abla}).  The best result in each case is shown in bold.} 
\vspace{-2ex}
\label{tab:coherence_check}
\end{table}

%% file: figs/pipeline/pipeline.tex
\begin{figure}[t]
    \centering
    \includegraphics[width=1.0\linewidth]{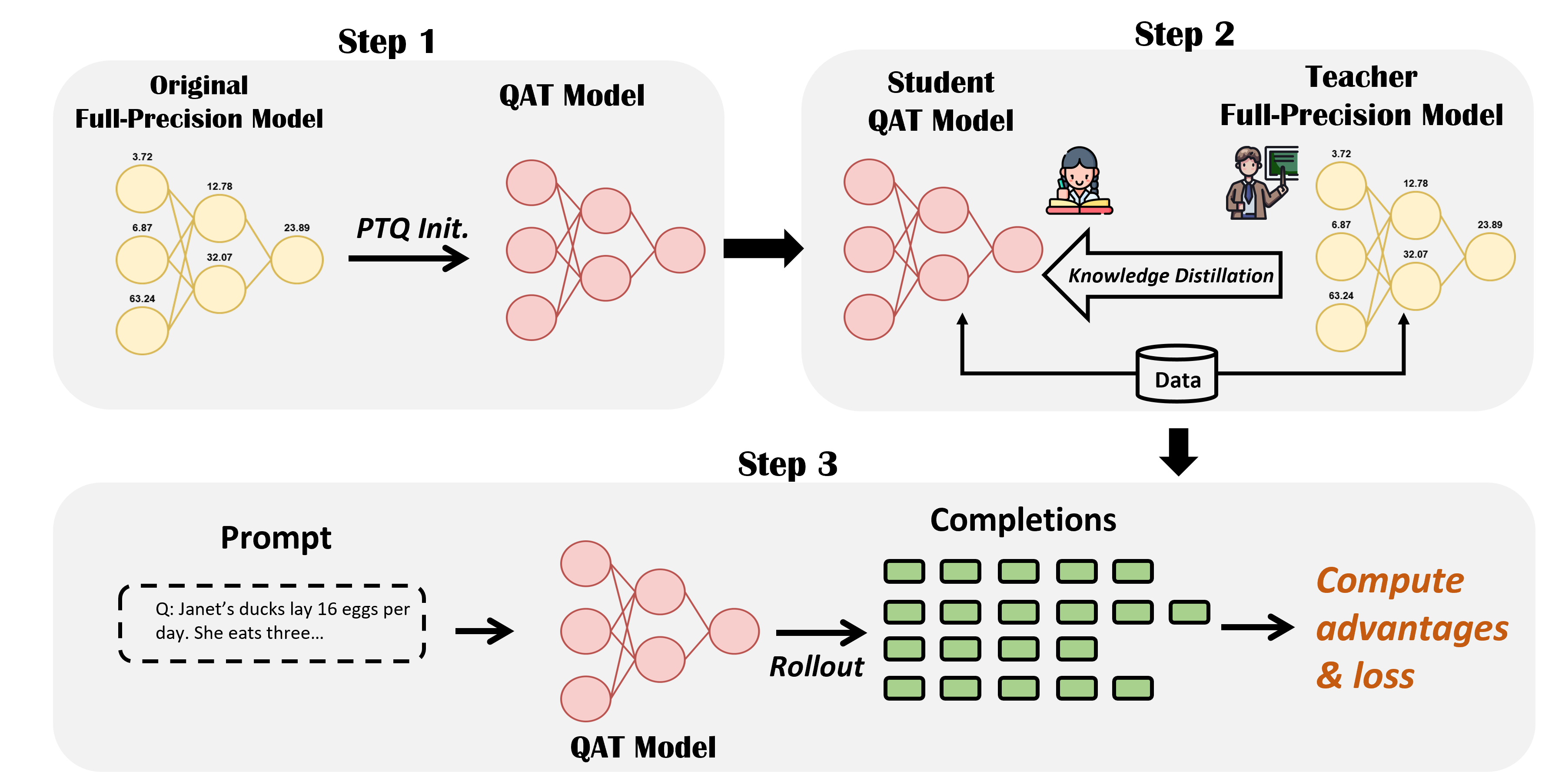}
    \caption{The overall workflow of the Reasoning-QAT. Step~1: PTQ-based initialization can provide a better starting point. Step~2: KD from the original full-precision model to align the teacher's behavior, and also serve as a cold-start for subsequent RL. Step~3: Based on cold-start, RL can further recover the reasoning ability of the QAT model. }
    \label{fig:pipeline}
\end{figure}

%% file: sections/conclusion.tex
\section{Conclusion}
We present a systematic empirical study of QAT for reasoning models under extreme low-bit quantization. We identify four key findings: KD is the preferred recovery objective, PTQ initialization improves training efficiency and accuracy, RL provides additional gains when a viable cold start exists (with KD acting as a practical prerequisite), and PTQ--QAT domain alignment affects convergence and accuracy. Finally, we consolidate these findings into Reasoning-QAT, a three-stage workflow that validates our findings and achieves a strong recovery compared to PTQ baselines.

\section*{Limitations}
First, our QAT training utilizes primarily math-centric data. While this effectively recovers performance on mathematical benchmarks, cross-domain generalization to other reasoning tasks (e.g., coding and science) remains limited, particularly for smaller models under extreme low-bit settings. Second, as an empirical study, this work focuses on validating optimal training workflows rather than designing new quantization methods.

%% file: sections/appendix.tex
\section{Appendix}
\subsection{Training Implementations Details}
\label{Implementations Details}
We list the detailed training hyperparameters in Tables \ref{tab:phase1_details} and \ref{tab:phase2_rl_details}. 

\label{appendix:train_details}

\begin{table*}[htbp]
\centering
\begin{minipage}[t]{0.58\textwidth}
    \vspace{0pt}
    \centering
    \scriptsize
    \setlength{\tabcolsep}{3.5pt} 
    \renewcommand{\arraystretch}{1.1}

    \begin{tabular}{l cccc c}
        \toprule
        \multirow{3}{*}{\textbf{Parameter}} & \multicolumn{4}{c}{\textbf{W3G128}} & \textbf{W2G128} \\
        \cmidrule(lr){2-5} \cmidrule(lr){6-6}
        & \multicolumn{2}{c}{\textbf{RTN Init}} & \multicolumn{2}{c}{\textbf{GPTQ Init}} & \textbf{GPTQ Init.} \\
        \cmidrule(lr){2-3} \cmidrule(lr){4-5} \cmidrule(lr){6-6}
        & \textbf{SFT} & \textbf{KD} & \textbf{SFT} & \textbf{KD} & \textbf{KD} \\
        \midrule
        \multicolumn{6}{l}{\textit{\textbf{Optimizer}}} \\
        Optimizer & \multicolumn{5}{c}{Adam} \\
        LR (Peak) & 2e-5 & 2e-5 & 1e-6 & 1e-6 & 5e-5 \\
        LR Scheduler & \multicolumn{5}{c}{Cosine Decay} \\
        Warmup Steps & 180 & 180 & 180 & 180 & 180 \\
        Adam Betas ($\beta_1, \beta_2$) & \multicolumn{5}{c}{0.9, 0.95} \\
        \midrule
        \multicolumn{6}{l}{\textit{\textbf{Training}}} \\
        Global Batch Size & \multicolumn{5}{c}{32} \\
        Gradient Accumulation & \multicolumn{5}{c}{4} \\
        Training Steps & 6,000 & 6,000 & 6,000 & 6,000 & 6,000 \\
        \bottomrule
    \end{tabular}
    \caption{Hyperparameters for Phase 1 (Cold Start).}
    \label{tab:phase1_details}
\end{minipage}
\hspace{0.5cm} 
\begin{minipage}[t]{0.28\textwidth}
    \vspace{0pt}
    \centering
    \scriptsize
    \setlength{\tabcolsep}{5pt} 
    \renewcommand{\arraystretch}{1.1}

    \begin{tabular}{lc}
        \toprule
        \textbf{Parameter} & \textbf{Value} \\
        \midrule
        \multicolumn{2}{l}{\textit{\textbf{Optimizer}}} \\
        Optimizer & Adam \\
        LR (Peak) & 5e-7 \\
        Scheduler & Cosine \\
        Warmup & 8 \\
        Betas & 0.9, 0.95 \\
        \midrule
        \multicolumn{2}{l}{\textit{\textbf{Training}}} \\
        Batch Size & 64 \\
        Grad Acc & 4 \\
        Steps & 250 \\
        \midrule
        \multicolumn{2}{l}{\textit{\textbf{Rollout}}} \\
        Group Size & 8 \\
        Max Len & 32768 \\
        \bottomrule
    \end{tabular}
    \caption{Hyperparameters for Phase 2 (RL).}
    \label{tab:phase2_rl_details}
\end{minipage}
\end{table*}

\subsection{Complete Factorial Combinations on Weight-Only Settings}
\label{complete_combination}

\begin{table*}[t] 
 \centering 
 \resizebox{0.9\linewidth}{!}{%
 \begin{tabular}{c|ccccc|ccc|c} 
 \toprule 
 & \textbf{RTN} & \textbf{GPTQ} & \textbf{SFT} & \textbf{KD} & \textbf{GRPO} & \textbf{AIME120} & \textbf{MATH-500} & \textbf{GSM8K} & \textbf{AVG} \\ 
\cline{1-10}
 \#0& - & - & - & - & - & 21.67 & 84.40 & 84.61 & 63.56 \\ 
 \cline{1-10}
 \#1& \checkmark & - & - & - &  & 0.83 & 15.00 & 15.39 & 10.41 \\ 
 \#2& \checkmark & - & \checkmark & - & - & 10.00 & 73.60 & 75.54 & 53.05 \\ 
 \#3& \checkmark & - & - & \checkmark & - & 14.44 & 76.20 & 75.87 & 55.50 \\
 \#4& \checkmark & - & - & \checkmark & \checkmark & 14.44 & 78.00 & 77.93 & 56.79 \\ 
 \cline{1-10} 
 \#5& - & \checkmark & - & - & - & 10.00 & 71.60 & 75.66 & 52.42 \\ 
 \#6& - & \checkmark & \checkmark & - & - & 14.17 & 75.53 & 76.12 & 55.27\\ 
 \#7& - & \checkmark & - & \checkmark & - & 13.89 & 78.20 & 77.26 & 56.45 \\ 
 \#8& - & \checkmark & - & \checkmark & \checkmark & \textbf{16.39} & \textbf{79.80} & \textbf{79.35} & \textbf{58.51}\\ 
 \bottomrule 
 \end{tabular}%
 } 
 \caption{Ablation studies of Reasoning-QAT, including the PTQ initializations~(i.e., RTN and GPTQ), QAT training paradigms~(i.e., SFT, KD and GRPO) based on R1-Qwen-1.5B. }
 \label{tab:final_abla} 
 \end{table*}

In this section, we clarify the efficacy of each Reasoning-QAT components, which are PTQ initialization, KD, and GRPO. We specifically assess the 3-bit groupwise weight-only quantization on R1-Qwen-1.5B model shown in Table~\ref{tab:final_abla}. 

\paragraph{The Effect of PTQ Initialization.} To investigate the impact of different weight quantization initialization strategies on the effectiveness of QAT, we present QAT models starting from RTN and GPTQ in rows 1-4 and rows 5-8, respectively. It can be found that using GPTQ for initialization yields a better starting point, resulting in an average improvement of 42.01\% (row 1 vs. row 5), which is also consistent with the findings in Section~\ref{sec:rq2}.

\paragraph{The Effect of KD.} Both SFT and KD significantly recover quantization loss. With RTN initialization, SFT yields a 42.64\% improvement (row 1 vs. row 2), while KD achieves a 45.09\% gain (row 1 vs. row3). Regardless of initialization, the KD approach demonstrates robustly superior performance over SFT. To be specific, KD achieves higher average accuracy than SFT by 2.45\% under RTN (row 1 vs. row 2) and by 1.18\% under GPTQ (row 3 vs. row 4).

\paragraph{Further Improvement by GRPO.} To further refine the performance of quantized models trained with knowledge distillation (KD), we integrate GRPO into the training pipeline. It can be seen that GRPO further boosts KD performance by 1.29\% under RTN (row 3 vs. row 4) and 2.06\% under GPTQ (row 7 vs. row 8), demonstrating its effectiveness in enhancing quantized models through policy refinement. In particular, compared with RTN, GPTQ also provides a better starting point for the RL training.

\subsection{Extensive Experiments on Weight-Activation Quantization}
\label{extensive_wa_exp}
We then examine  W4A4KV4 quantization as a representative configuration for weight-activation quantization. This scenario is particularly challenging since weights, activations, and KV cache are quantized to low bits. Note that we implement Reasoning-QAT in this setting by loading the transformation matrices from FlatQuant as initialization and further performing QAT. Unlike the original FlatQuant, which applies layer-wise correction in isolation, our method uses network-wise adjustments during QAT. This holistic optimization makes the model account for the propagation of quantization errors across layers, thereby handling the accumulation of mismatches that single-layer correction cannot capture. As a result, the model can adaptively correct quantization errors in a globally consistent manner rather than relying solely on static PTQ calibration.

\paragraph{Experiment Settings.}
To further validate our method, we evaluate a joint 4-bit weight and 4-bit activation (W4A4) quantization setting. Following the FlatQuant framework, we employ per-channel symmetric quantization for weights and per-token asymmetric quantization for activations. During the KD stage, the model is fine-tuned on 48,000 sequences from the OpenR1-Math dataset, with a fixed sequence length of 8,192. To optimize training stability, we adopt a multi-tier learning rate strategy: standard weights are updated at $1 \times 10^{-6}$, while the transformation matrices and clipping factors in FlatQuant are assigned higher rates of $5 \times 10^{-5}$ and $5 \times 10^{-4}$, respectively. Configuration for the ensuing Reinforcement Learning (RL) phase remains consistent with the details provided in \ref{tab:phase2_rl_details}.

\paragraph{Comparison with Representative PTQ Methods} 

\begin{table*}[t]
\centering
\resizebox{1.0\linewidth}{!}{%
\begin{tabular}{p{2 cm}|c|c|rrrrr|r|r}
    \toprule
    \textbf{Model} & \textbf{W-Bits} &  \textbf{Methods} & \textbf{AIME-120} & \textbf{MATH-500} & \textbf{GSM8K} & \textbf{\makecell{GPQA-\\Diamond}} & \textbf{\makecell{LiveCode-\\Bench}} & \textbf{Avg.} & \textbf{Drop $\downarrow$} \\
    \midrule
    \multirowcell{4}{\centering{Qwen3-0.6B}}
    & BF16 &  - & 11.11 & 74.00 & 79.00 & 28.45 & 12.94 & 41.10 & - \\
    \cline{2-10}
    & \multirowcell{3}{W4A4KV4}
    &  QuaRot & 0.00 & 0.00 & 0.00 & 24.24 & 0.00 & 4.84 & -36.26 \\
    & &  FlatQuant & 0.28 & 21.67 & 33.06 & 29.80 & 1.87 & 17.34 & -23.76 \\
    & & \cellcolor{red!10} Reasoning-QAT & \cellcolor{red!10} 0.00 & \cellcolor{red!10} 30.27 & \cellcolor{red!10} 48.62 & \cellcolor{red!10} 26.94 &\cellcolor{red!10} 1.37 & \cellcolor{red!10} 21.44 &\cellcolor{red!10} -19.66 \\
    \cline{1-10}

    \multirowcell{4}{\centering{R1-1.5B}}
    & BF16 &  - & 21.67 & 84.40 & 84.61 & 36.87 & 16.04 & 48.72 & - \\
    \cline{2-10}
    & \multirowcell{3}{W4A4KV4}
    &  QuaRot & 0.00 & 1.20 & 0.76 & 8.59 & 0.00 & 2.11 & -46.61 \\
    & &  FlatQuant & 10.00 & 64.80 & 78.62 & 31.82 & 6.72 & 38.39 & -10.33 \\
    & & \cellcolor{red!10} Reasoning-QAT & \cellcolor{red!10} 12.50 &\cellcolor{red!10} 73.20 &\cellcolor{red!10} 77.94 &\cellcolor{red!10} 32.83 &\cellcolor{red!10} 10.07 &\cellcolor{red!10} 41.31 &\cellcolor{red!10} -7.41 \\
    \cline{1-10}   
    \multirowcell{3}{\centering{Qwen3-4B}}
    & BF16 &  - & 58.89 & 95.33 & 94.49 & 56.06 & 48.38 & 70.63 & - \\
    \cline{2-10}
    & \multirowcell{2}{W4A4KV4}
    &  FlatQuant & 32.78 & 89.93 & 92.12 & 47.47 & 29.10 & 58.28 & -12.35 \\
    & &\cellcolor{red!10}  Reasoning-QAT &\cellcolor{red!10} 36.67 & \cellcolor{red!10} 91.40 &\cellcolor{red!10} 92.42 &\cellcolor{red!10} 48.48 & \cellcolor{red!10} 34.95 &\cellcolor{red!10} 60.78 &\cellcolor{red!10} -9.85 \\
    \bottomrule
\end{tabular}%
}
\caption{Main results of Reasoning-QAT on Qwen3-0.6B, R1-Qwen-1.5B and Qwen3-4B across various reasoning benchmarks. }
\label{app:wa_results}
\end{table*}

As shown in ~\ref{app:wa_results}, PTQ baselines such as QuaRot ~\citep{ashkboos2024quarot} and FlatQuant ~\citep{sun2024flatquant} suffer from large performance decreases. Our workflow, however, achieves consistent improvements across all model sizes. For instance, on Qwen3-4B, Reasoning-QAT raises the average score from 58.28\% (FlatQuant) to 60.78\%, effectively narrowing the gap to full precision and demonstrating that our consolidated workflow can effectively tackle the degradation in W4A4KV4 quantization scenarios and shows the effectiveness of our four key findings. Note that QuaRot results for Qwen3-4B are omitted because its hidden size of 2560 is incompatible with the Hadamard transformation required by the standard kernel.

\subsection{Additional Evaluation on General Domains}
\label{app:general_tasks_w4}

\paragraph{Performance on Weight-Only Setting.}
To verify whether our reasoning-focused QAT workflow compromises the model's general capabilities, we evaluated Reasoning-QAT on non-reasoning benchmarks, including HellaSwag (Commonsense) ~\citep{zellers2019hellaswag}, PIQA (Physics)~\citep{bisk2020piqa}, and Winogrande (Commonsense)~\citep{sakaguchi2021winogrande}.
As shown in Table~\ref{tab:general_tasks_w3}, under the W3G128 setting, Reasoning-QAT maintains performance comparable to the PTQ baseline across all models and tasks, and even achieves slight improvements on PIQA (e.g., +2.45\% on R1-Qwen-1.5B).
This indicates that while our method is optimized for reasoning, it effectively recovers precision without catastrophic forgetting of general knowledge.
We observe similar stability in the W4A4KV4 setting, with detailed results provided in Table~\ref{tab:general_tasks_w4}.

\begin{table*}[t]
\centering
\small
\begin{tabular}{l l c c c}
\toprule
\textbf{Model} & \textbf{Method} & \textbf{HellaSwag} & \textbf{PIQA} & \textbf{Winogrande} \\
\midrule
\multirow{2}{*}{Qwen3-0.6B} & GPTQ (Baseline) & 29.35 & 59.36 & \textbf{50.28} \\
 & \textbf{Reasoning-QAT} & \textbf{29.41} & \textbf{59.79} & 50.19 \\
\midrule
\multirow{2}{*}{R1-Qwen-1.5B} & GPTQ (Baseline) & 32.85 & 60.01 & \textbf{51.22} \\
 & \textbf{Reasoning-QAT} & \textbf{33.96} & \textbf{62.46} & 50.71 \\
\midrule
\multirow{2}{*}{Qwen3-4B} & GPTQ (Baseline) & 42.83 & 69.10 & \textbf{53.07} \\
 & \textbf{Reasoning-QAT} & \textbf{42.95} & \textbf{70.13} & 53.00 \\
\bottomrule
\end{tabular}
\caption{Comparison of Reasoning-QAT and GPTQ on general domain benchmarks under W3G128 quantization. Our method maintains robust performance on non-reasoning tasks.}
\label{tab:general_tasks_w3}
\end{table*}

\begin{table*}[t]
\centering
\small

\begin{tabular}{l l c c c}
\toprule
\textbf{Model} & \textbf{Method} & \textbf{HellaSwag} & \textbf{PIQA} & \textbf{Winogrande} \\
\midrule
\multirow{2}{*}{Qwen3-0.6B} & FlatQuant (Baseline) & 34.39 & 62.24 & 52.05 \\
 & \textbf{Reasoning-QAT} & \textbf{34.42} & \textbf{62.40} & \textbf{52.13} \\
\midrule
\multirow{2}{*}{R1-Qwen-1.5B} & FlatQuant (Baseline) & 34.75 & 61.26 & \textbf{51.38} \\
 & \textbf{Reasoning-QAT} & \textbf{34.78} & \textbf{62.35} & 50.32 \\
\midrule
\multirow{2}{*}{Qwen3-4B} & FlatQuant (Baseline) & 44.94 & 69.10 & 54.10 \\
 & \textbf{Reasoning-QAT} & \textbf{46.03} & \textbf{70.35} & \textbf{54.30} \\
\bottomrule
\end{tabular}
\caption{Comparison of Reasoning-QAT and FlatQuant on general domain benchmarks under W4A4KV4 quantization.}
\label{tab:general_tasks_w4}
\end{table*}

\subsection{Analysis on Initialization for W4A4KV4 QAT}
\label{app:w4a4_init_analysis}

To address concerns regarding the generalizability of our approach and its dependency on specific PTQ methods (e.g., FlatQuant), we conducted additional studies on the W4A4KV4 setting using the R1-Qwen-1.5B model.

\paragraph{Transferability to Other PTQ Methods (QuaRot).}
We investigated whether Reasoning-QAT can transfer to other advanced PTQ methods, such as QuaRot~\citep{ashkboos2024quarot}.
As shown in Table~\ref{tab:quarot_ablation}, while the KD stage successfully transfers and improves performance over the PTQ baseline (e.g., MATH-500 improves from 1.20\% to 11.20\%), the subsequent RL stage collapses.
This empirically validates our insight in Section \ref{sec:rq3} that RL requires a minimum capability threshold to generate rewardable trajectories.
Since the QuaRot-based model (for this specific architecture) retained severe outliers even after KD, it failed to provide a viable starting policy for RL.
This confirms that while the QAT workflow is transferable, a strong initialization (like FlatQuant) is a prerequisite for the RL stage in W4A4 scenarios.

\begin{table*}[t]
\centering
\small

\resizebox{\textwidth}{!}{
\begin{tabular}{l l c c c c c c}
\toprule
\textbf{Initialization} & \textbf{Method} & \textbf{AIME-120} & \textbf{MATH-500} & \textbf{GSM8K} & \textbf{GPQA-Diamond} & \textbf{LCB} & \textbf{Status} \\
\midrule
\multirow{3}{*}{QuaRot} & PTQ & 0.00 & 1.20 & 0.76 & 8.59 & 0.00 & Collapsed \\
 & + KD & 1.67 & 11.20 & 12.74 & 9.09 & 1.12 & Improved \\
 & + KD + RL & 0.00 & 3.60 & 6.07 & 2.53 & 0.00 & Collapsed \\
\midrule
\multirow{2}{*}{FlatQuant} & PTQ & 10.00 & 64.80 & 78.62 & 31.82 & 6.72 & Converged \\
 & + KD + RL & \textbf{12.50} & \textbf{73.20} & \textbf{77.94} & \textbf{32.83} & \textbf{10.07} & \textbf{Converged} \\
\bottomrule
\end{tabular}
}
\caption{Investigating the transferability of Reasoning-QAT using QuaRot initialization on R1-Qwen-1.5B (W4A4KV4). While KD improves performance, the model fails to sustain RL training due to insufficient base capability.}
\label{tab:quarot_ablation}
\end{table*}

\paragraph{Necessity of Outlier Suppression (vs. RTN).}
We further justify the use of FlatQuant by comparing it with a standard RTN initialization (i.e., QAT without outlier suppression).
As presented in Table~\ref{tab:rtn_w4a4_collapse}, applying QAT directly on RTN initialization leads to complete training collapse.
This confirms that FlatQuant (or equivalent outlier-suppression techniques) is not a confounding variable but a \textit{necessary precondition} to condition the optimization landscape for stable W4A4 training.

\begin{table*}[!t]
\centering
\small

\resizebox{\textwidth}{!}{
\begin{tabular}{l l c c c c c c}
\toprule
\textbf{Initialization} & \textbf{Method} & \textbf{AIME-120} & \textbf{MATH-500} & \textbf{GSM8K} & \textbf{GPQA-Diamond} & \textbf{LCB} & \textbf{Status} \\
\midrule
\multirow{2}{*}{RTN} & KD & 0.00 & 0.60 & 1.29 & 4.55 & 0.00 & Collapsed \\
 & Reasoning-QAT & 0.00 & 0.20 & 0.37 & 1.01 & 0.00 & Collapsed \\
\midrule
FlatQuant & Reasoning-QAT & \textbf{12.50} & \textbf{73.20} & \textbf{77.94} & \textbf{32.83} & \textbf{10.07} & \textbf{Converged} \\
\bottomrule
\end{tabular}
}
\caption{Comparison of RTN vs. FlatQuant initialization for W4A4KV4 QAT on R1-Qwen-1.5B. Without outlier suppression (FlatQuant), QAT faces complete collapse.}
\label{tab:rtn_w4a4_collapse}
\end{table*}

\subsection{Artifacts and Licenses}
We utilize the following scientific artifacts, strictly adhering to their respective licenses and terms of use.

\paragraph{Models}
\begin{itemize}
    \item DeepSeek-R1-Distill-Qwen-1.5B: Released under the MIT License.
    \item Qwen3 Family (0.6B and 4B): Released under the Apache 2.0 License.
\end{itemize}

\paragraph{Datasets}
\begin{itemize}
    \item OpenR1-Math: Used for training and distillation, released under the Apache 2.0 License.
    \item Wikitext-2: Used for calibration comparison, released under the Creative Commons Attribution-ShareAlike (CC BY-SA) License.
    \item Evaluation Benchmarks: We use standard evaluation datasets including GSM8K, MATH-500, AIME-120, GPQA-Diamond, and LiveCodeBench. These are publicly available and used in accordance with their original licenses (typically MIT or Apache 2.0) for research evaluation purposes.
\end{itemize}

We confirm that all artifacts were used consistent with their intended use. We verified that the data subsets used do not contain personally identifying information (PII) or offensive content.

\subsection{Potential Risks and Ethical Considerations}
Our work focuses on the quantization of reasoning models to improve inference efficiency. While the quantized models generally retain the capabilities of their full-precision counterparts, they may inherit biases or hallucinations inherent in the base large language models (LLMs). Our methods do not introduce additional safety risks beyond those already present in the pre-trained backbones.

\subsection{AI Writing Assistance}
We utilized AI assistants strictly for grammatical error correction and text polishing. All scientific claims, experimental designs, and results presented in this paper were verified by the authors.

%% file: custom.bib
@article{zellers2019hellaswag,
  title={Hellaswag: Can a machine really finish your sentence?},
  author={Zellers, Rowan and Holtzman, Ari and Bisk, Yonatan and Farhadi, Ali and Choi, Yejin},
  journal={arXiv preprint arXiv:1905.07830},
  year={2019}
}

@inproceedings{bisk2020piqa,
  title={Piqa: Reasoning about physical commonsense in natural language},
  author={Bisk, Yonatan and Zellers, Rowan and Gao, Jianfeng and Choi, Yejin and others},
  booktitle={Proceedings of the AAAI conference on artificial intelligence},
  volume={34},
  number={05},
  pages={7432--7439},
  year={2020}
}

@article{sakaguchi2021winogrande,
  title={Winogrande: An adversarial winograd schema challenge at scale},
  author={Sakaguchi, Keisuke and Bras, Ronan Le and Bhagavatula, Chandra and Choi, Yejin},
  journal={Communications of the ACM},
  volume={64},
  number={9},
  pages={99--106},
  year={2021},
  publisher={ACM New York, NY, USA}
}

@article{sun2024flatquant,
  title={Flatquant: Flatness matters for llm quantization},
  author={Sun, Yuxuan and Liu, Ruikang and Bai, Haoli and Bao, Han and Zhao, Kang and Li, Yuening and Hu, Jiaxin and Yu, Xianzhi and Hou, Lu and Yuan, Chun and others},
  journal={arXiv preprint arXiv:2410.09426},
  year={2024}
}

@article{liu2025quantization,
  title={Quantization hurts reasoning? an empirical study on quantized reasoning models},
  author={Liu, Ruikang and Sun, Yuxuan and Zhang, Manyi and Bai, Haoli and Yu, Xianzhi and Yu, Tiezheng and Yuan, Chun and Hou, Lu},
  journal={arXiv preprint arXiv:2504.04823},
  year={2025}
}

@article{li2024evaluating,
  title={Evaluating quantized large language models},
  author={Li, Shiyao and Ning, Xuefei and Wang, Luning and Liu, Tengxuan and Shi, Xiangsheng and Yan, Shengen and Dai, Guohao and Yang, Huazhong and Wang, Yu},
  journal={arXiv preprint arXiv:2402.18158},
  year={2024}
}

@article{guo2025deepseek,
  title={Deepseek-r1: Incentivizing reasoning capability in llms via reinforcement learning},
  author={Guo, Daya and Yang, Dejian and Zhang, Haowei and Song, Junxiao and Zhang, Ruoyu and Xu, Runxin and Zhu, Qihao and Ma, Shirong and Wang, Peiyi and Bi, Xiao and others},
  journal={arXiv preprint arXiv:2501.12948},
  year={2025}
}

@misc{openr1,
    title = {Open R1: A fully open reproduction of DeepSeek-R1},
    url = {https://github.com/huggingface/open-r1},
    author = {Hugging Face},
    month = {January},
    year = {2025}
}

@inproceedings{rein2024gpqa,
  title={Gpqa: A graduate-level google-proof q\&a benchmark},
  author={Rein, David and Hou, Betty Li and Stickland, Asa Cooper and Petty, Jackson and Pang, Richard Yuanzhe and Dirani, Julien and Michael, Julian and Bowman, Samuel R},
  booktitle={First Conference on Language Modeling},
  year={2024}
}

@article{jain2024livecodebench,
  title={Livecodebench: Holistic and contamination free evaluation of large language models for code},
  author={Jain, Naman and Han, King and Gu, Alex and Li, Wen-Ding and Yan, Fanjia and Zhang, Tianjun and Wang, Sida and Solar-Lezama, Armando and Sen, Koushik and Stoica, Ion},
  journal={arXiv preprint arXiv:2403.07974},
  year={2024}
}

@article{cobbe2021gsm8k,
  title={Training Verifiers to Solve Math Word Problems},
  author={Cobbe, Karl and Kosaraju, Vineet and Bavarian, Mohammad and Chen, Mark and Jun, Heewoo and Kaiser, Lukasz and Plappert, Matthias and Tworek, Jerry and Hilton, Jacob and Nakano, Reiichiro and Hesse, Christopher and Schulman, John},
  journal={arXiv preprint arXiv:2110.14168},
  year={2021}
}

@misc{lighteval,
  author = {Fourrier, Clémentine and Habib, Nathan and Kydlíček, Hynek and Wolf, Thomas and Tunstall, Lewis},
  title = {LightEval: A lightweight framework for LLM evaluation},
  year = {2023},
  version = {0.7.0},
  url = {https://github.com/huggingface/lighteval}
}

@inproceedings{kwon2023efficient,
  title={Efficient memory management for large language model serving with pagedattention},
  author={Kwon, Woosuk and Li, Zhuohan and Zhuang, Siyuan and Sheng, Ying and Zheng, Lianmin and Yu, Cody Hao and Gonzalez, Joseph and Zhang, Hao and Stoica, Ion},
  booktitle={Proceedings of the 29th Symposium on Operating Systems Principles},
  pages={611--626},
  year={2023}
}

@inproceedings{lightman2023let,
  title={Let's verify step by step},
  author={Lightman, Hunter and Kosaraju, Vineet and Burda, Yuri and Edwards, Harrison and Baker, Bowen and Lee, Teddy and Leike, Jan and Schulman, John and Sutskever, Ilya and Cobbe, Karl},
  booktitle={The Twelfth International Conference on Learning Representations},
  year={2023}
}

@article{shao2024deepseekmath,
  title={Deepseekmath: Pushing the limits of mathematical reasoning in open language models},
  author={Shao, Zhihong and Wang, Peiyi and Zhu, Qihao and Xu, Runxin and Song, Junxiao and Bi, Xiao and Zhang, Haowei and Zhang, Mingchuan and Li, YK and Wu, Y and others},
  journal={arXiv preprint arXiv:2402.03300},
  year={2024}
}

@article{ashkboos2024quarot,
  title={Quarot: Outlier-free 4-bit inference in rotated llms},
  author={Ashkboos, Saleh and Mohtashami, Amirkeivan and Croci, Maximilian L and Li, Bo and Jaggi, Martin and Alistarh, Dan and Hoefler, Torsten and Hensman, James},
  journal={arXiv preprint arXiv:2404.00456},
  year={2024}
}

@inproceedings{frantar2022optq,
  title={OPTQ: Accurate quantization for generative pre-trained transformers},
  author={Frantar, Elias and Ashkboos, Saleh and Hoefler, Torsten and Alistarh, Dan},
  booktitle={The Eleventh International Conference on Learning Representations},
  year={2022}
}

@article{lin2023awq,
  title={Awq: Activation-aware weight quantization for llm compression and acceleration},
  author={Lin, Ji and Tang, Jiaming and Tang, Haotian and Yang, Shang and Chen, Wei-Ming and Wang, Wei-Chen and Xiao, Guangxuan and Dang, Xingyu and Gan, Chuang and Han, Song},
  journal={arXiv preprint arXiv:2306.00978},
  year={2023}
}

@article{liu2024intactkv,
  title={IntactKV: Improving Large Language Model Quantization by Keeping Pivot Tokens Intact},
  author={Liu, Ruikang and Bai, Haoli and Lin, Haokun and Li, Yuening and Gao, Han and Xu, Zhengzhuo and Hou, Lu and Yao, Jun and Yuan, Chun},
  journal={arXiv preprint arXiv:2403.01241},
  year={2024}
}

@article{ma2024affinequant,
  title={Affinequant: Affine transformation quantization for large language models},
  author={Ma, Yuexiao and Li, Huixia and Zheng, Xiawu and Ling, Feng and Xiao, Xuefeng and Wang, Rui and Wen, Shilei and Chao, Fei and Ji, Rongrong},
  journal={arXiv preprint arXiv:2403.12544},
  year={2024}
}

@article{lee2024improving,
  title={Improving conversational abilities of quantized large language models via direct preference alignment},
  author={Lee, Janghwan and Park, Seongmin and Hong, Sukjin and Kim, Minsoo and Chang, Du-Seong and Choi, Jungwook},
  journal={arXiv preprint arXiv:2407.03051},
  year={2024}
}

@article{chen2024efficientqat,
  title={Efficientqat: Efficient quantization-aware training for large language models},
  author={Chen, Mengzhao and Shao, Wenqi and Xu, Peng and Wang, Jiahao and Gao, Peng and Zhang, Kaipeng and Luo, Ping},
  journal={arXiv preprint arXiv:2407.11062},
  year={2024}
}

@article{liu2023llm,
  title={Llm-qat: Data-free quantization aware training for large language models},
  author={Liu, Zechun and Oguz, Barlas and Zhao, Changsheng and Chang, Ernie and Stock, Pierre and Mehdad, Yashar and Shi, Yangyang and Krishnamoorthi, Raghuraman and Chandra, Vikas},
  journal={arXiv preprint arXiv:2305.17888},
  year={2023}
}

@article{bondarenko2024low,
  title={Low-rank quantization-aware training for llms},
  author={Bondarenko, Yelysei and Del Chiaro, Riccardo and Nagel, Markus},
  journal={arXiv preprint arXiv:2406.06385},
  year={2024}
}

@article{tailor2020degree,
  title={Degree-quant: Quantization-aware training for graph neural networks},
  author={Tailor, Shyam A and Fernandez-Marques, Javier and Lane, Nicholas D},
  journal={arXiv preprint arXiv:2008.05000},
  year={2020}
}

@inproceedings{nagel2022overcoming,
  title={Overcoming oscillations in quantization-aware training},
  author={Nagel, Markus and Fournarakis, Marios and Bondarenko, Yelysei and Blankevoort, Tijmen},
  booktitle={International Conference on Machine Learning},
  pages={16318--16330},
  year={2022},
  organization={PMLR}
}

@article{liu2025paretoq,
  title={Paretoq: Scaling laws in extremely low-bit llm quantization},
  author={Liu, Zechun and Zhao, Changsheng and Huang, Hanxian and Chen, Sijia and Zhang, Jing and Zhao, Jiawei and Roy, Scott and Jin, Lisa and Xiong, Yunyang and Shi, Yangyang and others},
  journal={arXiv preprint arXiv:2502.02631},
  year={2025}
}

@article{team2025kimi,
  title={Kimi k2: Open agentic intelligence},
  author={Team, Kimi and Bai, Yifan and Bao, Yiping and Chen, Guanduo and Chen, Jiahao and Chen, Ningxin and Chen, Ruijue and Chen, Yanru and Chen, Yuankun and Chen, Yutian and others},
  journal={arXiv preprint arXiv:2507.20534},
  year={2025}
}

@article{yang2025qwen3,
  title={Qwen3 technical report},
  author={Yang, An and Li, Anfeng and Yang, Baosong and Zhang, Beichen and Hui, Binyuan and Zheng, Bo and Yu, Bowen and Gao, Chang and Huang, Chengen and Lv, Chenxu and others},
  journal={arXiv preprint arXiv:2505.09388},
  year={2025}
}

@article{hinton2015distilling,
  title={Distilling the knowledge in a neural network},
  author={Hinton, Geoffrey and Vinyals, Oriol and Dean, Jeff},
  journal={arXiv preprint arXiv:1503.02531},
  year={2015}
}

@article{du2024bitdistiller,
  title={Bitdistiller: Unleashing the potential of sub-4-bit llms via self-distillation},
  author={Du, Dayou and Zhang, Yijia and Cao, Shijie and Guo, Jiaqi and Cao, Ting and Chu, Xiaowen and Xu, Ningyi},
  journal={arXiv preprint arXiv:2402.10631},
  year={2024}
}

@article{frantar2022gptq,
  title={Gptq: Accurate post-training quantization for generative pre-trained transformers},
  author={Frantar, Elias and Ashkboos, Saleh and Hoefler, Torsten and Alistarh, Dan},
  journal={arXiv preprint arXiv:2210.17323},
  year={2022}
}

@article{jaech2024openai,
  title={Openai o1 system card},
  author={Jaech, Aaron and Kalai, Adam and Lerer, Adam and Richardson, Adam and El-Kishky, Ahmed and Low, Aiden and Helyar, Alec and Madry, Aleksander and Beutel, Alex and Carney, Alex and others},
  journal={arXiv preprint arXiv:2412.16720},
  year={2024}
}

@article{qu2025survey,
  title={A survey of efficient reasoning for large reasoning models: Language, multimodality, and beyond},
  author={Qu, Xiaoye and Li, Yafu and Su, Zhaochen and Sun, Weigao and Yan, Jianhao and Liu, Dongrui and Cui, Ganqu and Liu, Daizong and Liang, Shuxian and He, Junxian and others},
  journal={arXiv preprint arXiv:2503.21614},
  year={2025}
}

@article{li2025gptaq,
  title={GPTAQ: Efficient Finetuning-Free Quantization for Asymmetric Calibration},
  author={Li, Yuhang and Yin, Ruokai and Lee, Donghyun and Xiao, Shiting and Panda, Priyadarshini},
  journal={arXiv preprint arXiv:2504.02692},
  year={2025}
}

@article{li2025quantization,
  title={Quantization meets reasoning: Exploring llm low-bit quantization degradation for mathematical reasoning},
  author={Li, Zhen and Su, Yupeng and Yang, Runming and Xie, Congkai and Wang, Zheng and Xie, Zhongwei and Wong, Ngai and Yang, Hongxia},
  journal={arXiv preprint arXiv:2501.03035},
  year={2025}
}

@article{srivastava2025towards,
  title={Towards reasoning ability of small language models},
  author={Srivastava, Gaurav and Cao, Shuxiang and Wang, Xuan},
  journal={arXiv preprint arXiv:2502.11569},
  year={2025}
}

@article{wang2025bitnet,
  title={BitNet v2: Native 4-bit Activations with Hadamard Transformation for 1-bit LLMs},
  author={Wang, Hongyu and Ma, Shuming and Wei, Furu},
  journal={arXiv preprint arXiv:2504.18415},
  year={2025}
}

@article{jeon2024l4q,
  title={L4q: Parameter efficient quantization-aware training on large language models via lora-wise lsq},
  author={Jeon, Hyesung and Kim, Yulhwa and Kim, Jae-joon},
  journal={CoRR},
  year={2024}
}

@article{qin2024accurate,
  title={Accurate lora-finetuning quantization of llms via information retention},
  author={Qin, Haotong and Ma, Xudong and Zheng, Xingyu and Li, Xiaoyang and Zhang, Yang and Liu, Shouda and Luo, Jie and Liu, Xianglong and Magno, Michele},
  journal={arXiv preprint arXiv:2402.05445},
  year={2024}
}

@article{lin2024duquant,
  title={Duquant: Distributing outliers via dual transformation makes stronger quantized llms},
  author={Lin, Haokun and Xu, Haobo and Wu, Yichen and Cui, Jingzhi and Zhang, Yingtao and Mou, Linzhan and Song, Linqi and Sun, Zhenan and Wei, Ying},
  journal={Advances in Neural Information Processing Systems},
  volume={37},
  pages={87766--87800},
  year={2024}
}

@article{liu2025flexquant,
  title={FlexQuant: A Flexible and Efficient Dynamic Precision Switching Framework for LLM Quantization},
  author={Liu, Fangxin and Wang, Zongwu and Xia, JinHong and Zhao, Junping and Liu, Jian and Guan, Haibing and Jiang, Li},
  journal={arXiv preprint arXiv:2506.12024},
  year={2025}
}

@article{li2025kvtuner,
  title={KVTuner: Sensitivity-Aware Layer-Wise Mixed-Precision KV Cache Quantization for Efficient and Nearly Lossless LLM Inference},
  author={Li, Xing and Xing, Zeyu and Li, Yiming and Qu, Linping and Zhen, Hui-Ling and Liu, Wulong and Yao, Yiwu and Pan, Sinno Jialin and Yuan, Mingxuan},
  journal={arXiv preprint arXiv:2502.04420},
  year={2025}
}

@inproceedings{yangattentionpredictor,
  title={AttentionPredictor: Temporal Patterns Matter for KV Cache Compression},
  author={Yang, Qingyue and Wang, Jie and Li, Xing and Wang, Zhihai and Chen, Chen and Chen, Lei and Yu, Xianzhi and Liu, Wulong and Hao, Jianye and Yuan, Mingxuan and others},
  booktitle={The Thirty-ninth Annual Conference on Neural Information Processing Systems}
}

@article{zhang2026benchmarking,
  title={Benchmarking Post-Training Quantization of Large Language Models under Microscaling Floating Point Formats},
  author={Zhang, Manyi and Li, Ji-Fu and Sun, Zhongao and Bai, Haoli and Zhen, Hui-Ling and Dong, Zhenhua and Yu, Xianzhi},
  journal={arXiv preprint arXiv:2601.09555},
  year={2026}
}
